\documentclass[lettersize,journal]{IEEEtran}
\usepackage{amsmath,amsfonts}
\usepackage{array}
\usepackage[caption=false,font=normalsize,labelfont=sf,textfont=sf]{subfig}
\usepackage{textcomp}
\usepackage{stfloats}
\usepackage{url}
\usepackage{verbatim}
\usepackage{graphicx}
\usepackage[ruled,linesnumbered]{algorithm2e}
\usepackage{algorithmicx,algpseudocode}
\usepackage{algorithm2e}
\usepackage{cite}
\usepackage[pagebackref,breaklinks,colorlinks]{hyperref}
\usepackage{amssymb}
\usepackage{mathrsfs}
\usepackage{booktabs}
\usepackage{multirow}
\usepackage{subfig}
\usepackage{makecell}
\usepackage{bm}
\usepackage{bbding}
\usepackage{diagbox}
\usepackage[T1]{fontenc}
\usepackage{multicol}
\usepackage{lipsum} 
\usepackage{wrapfig}
\usepackage{soul} 
\usepackage{color, xcolor} 

\soulregister\cite7
\soulregister\ref7 
\soulregister\szequation7

\makeatletter
\renewcommand{\maketag@@@}[1]{\hbox{\m@th\normalsize\normalfont#1}}%
\makeatother
\newcommand{\szequation}[1]{{\normalsize #1}}
\renewcommand{\checkmark}{\Checkmark}
\begin{document}

\title{Pedestrian Trajectory Prediction Based on Social Interactions Learning with Random Weights}

\author{Jiajia Xie,
Sheng Zhang,~
Beihao Xia,~
~Zhu Xiao,~\IEEEmembership{Senior Member, IEEE},~
Hongbo Jiang,~\IEEEmembership{Senior Member, IEEE},~
Siwang~Zhou,~Zheng~Qin,
Hongyang Chen,~\IEEEmembership{Senior Member, IEEE}~

\thanks{Manuscript received 9 March 2023; revised 11 September 2023 and 3 Jan
uary 2024; accepted 12 February 2024. Date of publication 8 March 2024;
 date of current version 24 April 2024. This work was supported in part by the NSFC under Grants 62272152 and 62271452, in part by the National Key R\&D Program of China under Grants 2023YFC3321601 and 2022YFB4500300, in part by the Key R\&D Program of
Hunan Province under Grant 2022GK2020,~in part by the Shenzhen Science and Technology
Program under Grant JCYJ20220530160408019, in part by the CAAI-Huawei MindSpore Open Fund, and in part by the Guangdong Basic and Applied Basic Research Foundation under Grant 2023A1515011915. (\textit{Corresponding
authors:~B.~Xia, Z.~Xiao and H.~Chen.})}

\thanks{J.~Xie,~Z. Xiao,~H.~Jiang,~S.~Zhou~and~Z.~Qin are with the College of Computer Science and Electronic Engineering, Hunan University, Changsha, 410082, China, and also with the Shenzhen Research Institute, Hunan University, Shenzhen 518055. e-mail:~\{xiejiajia, zhxiao,~swzhou,~zqin\}@hnu.edu.cn,~hongbojiang2004@gmail.com.}

\thanks{S.~Zhang is with the School of Computer Science and Technology, Zhejiang University,~HangZhou, 310000,~China.~e-mail:~ghz@zju.edu.cn.}

\thanks{B.~Xia is with the School of Electronic Information and Communication, Huazhong University of Science and Technology,~Wu Han, 430074,~China.~e-mail:~xbh\_hust@hust.edu.cn.}

\thanks{H.~Chen is with Zhejiang Lab, Hangzhou, 311121,~China.~e-mail:~hongyang@zhejianglab.com.}

}




\maketitle

\begin{abstract}
Pedestrian trajectory prediction is a critical technology in the evolution of self-driving cars toward complete artificial intelligence.
Over recent years, focusing on the trajectories of pedestrians to model their social interactions has surged with great interest in more accurate trajectory predictions.
However, existing methods for modeling pedestrian social interactions rely on pre-defined rules,  struggling to capture non-explicit social interactions.
In this work, we propose a novel framework named DTGAN, which extends the application of Generative Adversarial Networks (GANs) to graph sequence data, with the primary objective of automatically capturing implicit social interactions and achieving precise predictions of pedestrian trajectory.
DTGAN innovatively incorporates random weights within each graph to eliminate the need for pre-defined interaction rules.
We further enhance the performance of DTGAN by exploring diverse task loss functions during adversarial training, which yields improvements of 16.7\% and 39.3\% on metrics ADE and FDE, respectively.
The effectiveness and accuracy of our framework are verified on two public datasets.
The experimental results show that our proposed DTGAN achieves superior performance and is well able to understand pedestrians' intentions.

\end{abstract}


\section{Introduction}

\IEEEPARstart{W}{}ith the rapid advancement of technology, there is a growing interest among researchers~\cite{XIAO1, 20TITS-vincent, XIAO5} in investigating potential patterns through the analysis of trajectories~\cite{XIAO2, XIAO3, XIAO4}.
Pedestrian trajectory prediction has raised considerable concern in recent years. It is the task of using observations from a past period to predict the likely trajectory for a future period.
In the context of autonomous driving~\cite{zhang2022trajectory, wong2022view, saleh2018intent, xia2022cscnet} and smart mobility~\cite{monfort2015intent,long2022location}, vehicles are required to predict the trajectories of nearby pedestrians to avoid collisions~\cite{23jsac, 23TMC}. 
However, the movement of pedestrians is highly uncertain~\cite{li2018deep}. To be more specific, their destinations, and even their intentions, often remain unknown. Moreover, pedestrians often rely on the movements of others to adjust their paths, such as following others, dispersing from a specific point, merging from different directions, walking in groups, and so on. 
Consequently, predicting pedestrian trajectories under such complex social interactions poses significant challenges.
\par

\begin{figure}[t]
\centering
\includegraphics[scale=0.32]{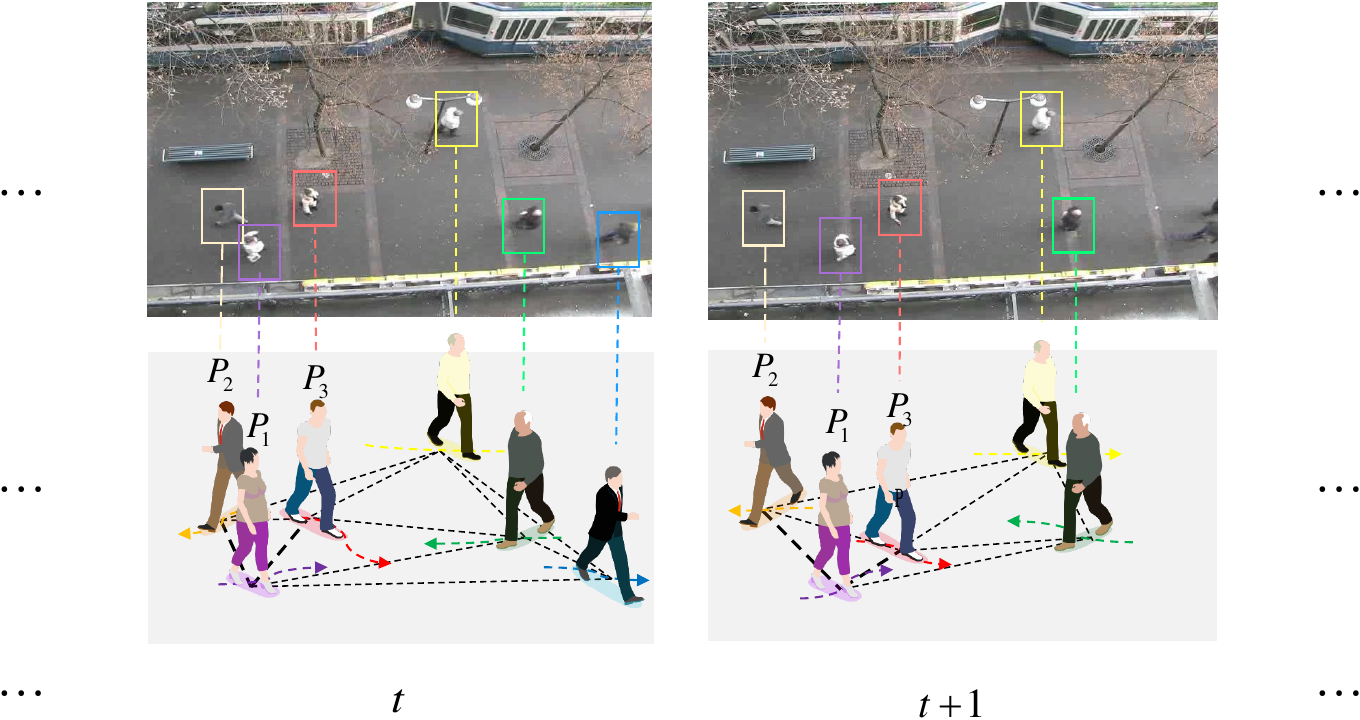}
\caption{{\small Pedestrian interactions graph representation. We model pedestrian trajectories at each time $t$ as graph-structured data, using different colored nodes to represent different pedestrians and edges to represent social interactions among pedestrians.}}
\label{fig:graph_pic}
\end{figure}

For seeking solutions, many works aim to model social interactions among pedestrians to increase the accuracy of trajectory prediction.
One such work is Social-LSTM \cite{alahi2016social}, which models social interactions through the ``Social pooling'' layer, sharing hidden states between neighboring LSTMs in space, yielding satisfactory results.
To eliminate such local neighborhood assumptions, Social Attention \cite{vemula2018social} employs a soft attention mechanism to capture social interactions for all pedestrians on the scene.
CIDNN \cite{xu2018encoding} uses spatial affinity to measure the influence of other pedestrians on the target pedestrian and accordingly weights their motion features.
Later works, like Social-GAN \cite{gupta2018social}, utilize a ``Max-Pooling'' mechanism to simulate all pedestrian interactions by selecting to pool hidden states of LSTMs at specific time steps. 
SoPhie \cite{sadeghian2019SoPhie} and Social-BiGAT \cite{kosaraju2019social} introduce an attention mechanism to improve the modeling of social interactions.  
SoPhie adopts two separate soft attention modules ordered by the Euclidean distance between the target and others, maintaining the uniqueness of the input compared to ``Max-Pooling''. Meanwhile, Social-BiGAT uses graph attention networks (GAT) to model both local and global interactions without pooling or sorting.
STGAT  \cite{huang2019stgat} encodes the spatial and temporal interactions and then leverages GAT to aggregate their feature information. 
Although the way in which social interactions among pedestrians are modeled has a substantial influence on trajectory prediction, interpreting the physical meaning of the recurrent unit states can be challenging and not intuitive.
For this reason, Social-STGCNN\cite{mohamed2020social} utilizes graph-based techniques to model social interactions among pedestrians. By weighting edges of nodes based on human-defined rules, STGCNN provides a clear, explicit quantification of social interactions within each time step.
The human-defined weighting system, while efficient and straightforward, remains static throughout the training epochs. This static representation may not fully encapsulate the dynamic and multi-aspects of real-world social interactions, which are influenced by various factors beyond mere physical distance.
Despite promoting results, such pre-defined weights might be insufficient to capture social interactions due to the potential for bias and incompleteness.
As shown in Fig. \ref{fig:graph_pic}, assume that social interactions are determined by Euclidean distance between pedestrians. We observe the target $P_1$ pays more attention to the pedestrian $P_2$ instead of pedestrian $P_3$, with whom they may collide. The situation remains that we pay more attention to people in front of us than the ones behind us when we walk. The use of pre-defined weights in neural networks hinders the extraction of implicit information from social interactions. The pre-defined weights limit the ability of neural networks to extract implicit information from social interactions. Therefore, more adaptive and data-driven methods are required to model social interactions.
\par

In this work, we propose a novel framework, \textit{Dynamic aTtention Generator Adversarial Networks} (DTGAN), which addresses the limitations of pre-defined weights by using graph sequence data with random weights to capture non-explicit social interactions.
By employing attention mechanisms and adversarial training, the framework can achieve improved accuracy in predicting diverse and realistic trajectories, considering multiple alternative paths to reach the pedestrians' destinations.
Firstly, random weights formulated as a randomly generated matrix replace the pre-defined matrix in the graph to eliminate potential bias, providing a more adaptive approach to modeling social interactions.
Secondly, DTGAN leverages the power of GAN \cite{mirza2014conditional,liu2022foreseeing,zhou2022recognition,9851855,li2022comprehensive,li2022fakeclr,li2020systematic} to generate multi-modal trajectories for pedestrians, considering diverse paths to address uncertainty in pedestrian movement. 
Specifically, given a sequence of random weight graphs, DTGAN utilizes GAT to assign different adaptive importance for nodes in each graph and incorporates adversarial training to improve its performance, extending the application of GAN to graph sequence data.
On top of that, DTGAN explores various task loss functions to encourage diverse and realistic trajectory predictions while maintaining alignment with ground truth, striking a balance between generating diverse trajectories and adhering to actual pedestrian behavior.
To obtain a good grasp of the properties produced by DTGAN, several experiments are conducted, including component assessment, task loss exploration, and the impact of random weight generation.
\par

The contributions of our work are highlighted as follows:
\begin{enumerate}
    \item  We propose a novel framework named DTGAN, which extends the application of GAN to graph sequence data with random weights, 
    eliminating potential bias and enhancing generalization to different scenes or datasets.
    \item  We improve the performance of the DTGAN by exploring different task loss functions in adversarial training, which is expected to encourage new directions of thinking about the GANs. 
    \item The experimental results demonstrate our framework achieves superior performance and effectively understands pedestrian intent, which further validates the promise of leveraging graphs with random weights.
\end{enumerate}

The remainder of this paper is organized as follows. Section \ref{section:relate work} reviews the related work. Section \ref{section:method} introduces details of the proposed model and different loss functions. Section \ref{section:Experiments} presents the experimental results, analyses, and limitations. Finally, Section \ref{section:conclusion} provides a summary of the work and an outlook for future work.

 \section{Related work}
 \label{section:relate work} 
\textbf{Human-human interactions.} ~
The Social Force method, as proposed by \cite{helbing1995social}, is a widely used model for simulating pedestrian motion by incorporating attractive and repulsive forces. While the attractive force guides pedestrians toward their intended destination, the repulsive force helps them avoid collisions with other pedestrians. The effectiveness of this method has led to its extension to various other fields, such as abnormal behavior detection \cite{mehran2009abnormal}, trajectory prediction \cite{leal2011everybody}, and crowd simulation \cite{luber2010people, pellegrini2010improving, feng2016social}.
In addition to the Social Force method, other models such as the Discrete Choice \cite{antonini2006discrete} and Gaussian processes utilized by \cite{wang2007Gaussian,tay2008modelling} have been explored. However, most of these models rely on hand-crafted energy potentials based on relative distances and specific rules.
\par

\textbf{GNN-based models.}~
STGAT \cite{huang2019stgat} models pedestrian motions by combining Graph Attention Networks (GATs) with Long Short-Term Memory (LSTM) in the context of pedestrian motions, which models the spatial interactions by the GAT-based module and captures the temporal correlations of interactions by the LSTM-based module. 
Although Social-BiGAT \cite{kosaraju2019social} also uses GATs to model social interactions, they both aggregate hidden states of LSTMs instead of acting on pedestrians directly.  However, this method may not capture the latest social interactions well when pedestrians change their intentions in a short period.
Graph Convolutional Networks (GCNs)\cite{kipf2017semi} is a method that extends the concept of Convolutional Neural Networks to graphs by performing direct convolution on the adjacency matrix of the graph. 
Inspired by this, Social-STGCNN \cite{mohamed2020social} quantifies the impact of interactions among pedestrians by constructing a distance-related function, using GCN instead of the previous theming mechanism, with a significant improvement in prediction performance. 
Other works \cite{yu2020spatio,li2022graph} extend the transformer network to graph structure data. 
Graph Neural Networks (GNNs) have shown great potential in modeling social interactions.
We follow previous works using graphs to simulate the social interactions and then apply a GNN-based model to extract interaction information.
\par

\textbf{GAN-based models.}  Social-LSTM\cite{alahi2016social} based on simple LSTM and Seq2Seq sequences, is among the pioneering works in the field and achieves satisfactory results in terms of prediction accuracy.
Social-LSTM models each individual's trajectory using an LSTM and introduces a social pooling layer to share their hidden state, capturing social interactions. 
Subsequent works extended the GAN framework \cite{li2019adversarial,USENIX_LI} to improve trajectory prediction. Social-GAN \cite{gupta2018social}, building upon Social-LSTM, combines LSTM with GAN and designs a ``Max-Pooling'' mechanism to aggregate pedestrian information, leading to diverse predictions of trajectories.
Social-BiGAT \cite{kosaraju2019social} introduces attention networks within LSTM-based GAN by utilizing images as additional input features. It generates multi-modal trajectories by using GAT and self-attention on images, aiming to capture both social interactions and scene context information.
This idea is akin to SoPhie \cite{sadeghian2019SoPhie}, which also uses images to enrich predictions by combining scene context and social interactions through LSTM and GAN. Motivated by the success of these GAN-based models, our work draws on the adversarial idea in GAN to generate multi-modal trajectories of pedestrians, which corresponds with the highly uncertain movement of pedestrians.
\par

Specifically, we incorporate random weights as edge weights in each graph to simulate the non-explicit interactions among pedestrians. These weights are not fixed but are aligned with the model's training, allowing the representation of social interactions to evolve dynamically with training epochs.
This practice aims to leverage the neural network to automatically capture the complex and dynamic nature of social interactions. We utilize the generative adversarial process to train our method. Since GAN can implicitly learn data distribution without the need for any supervisory information, which is consistent with our goal of avoiding pre-defined weights for graphs. By fusing graphs with random weights and GAN, we are able to address pedestrian trajectory prediction in a more significant manner.

\begin{figure*}[t]
    \centering
    \includegraphics[scale=0.49]{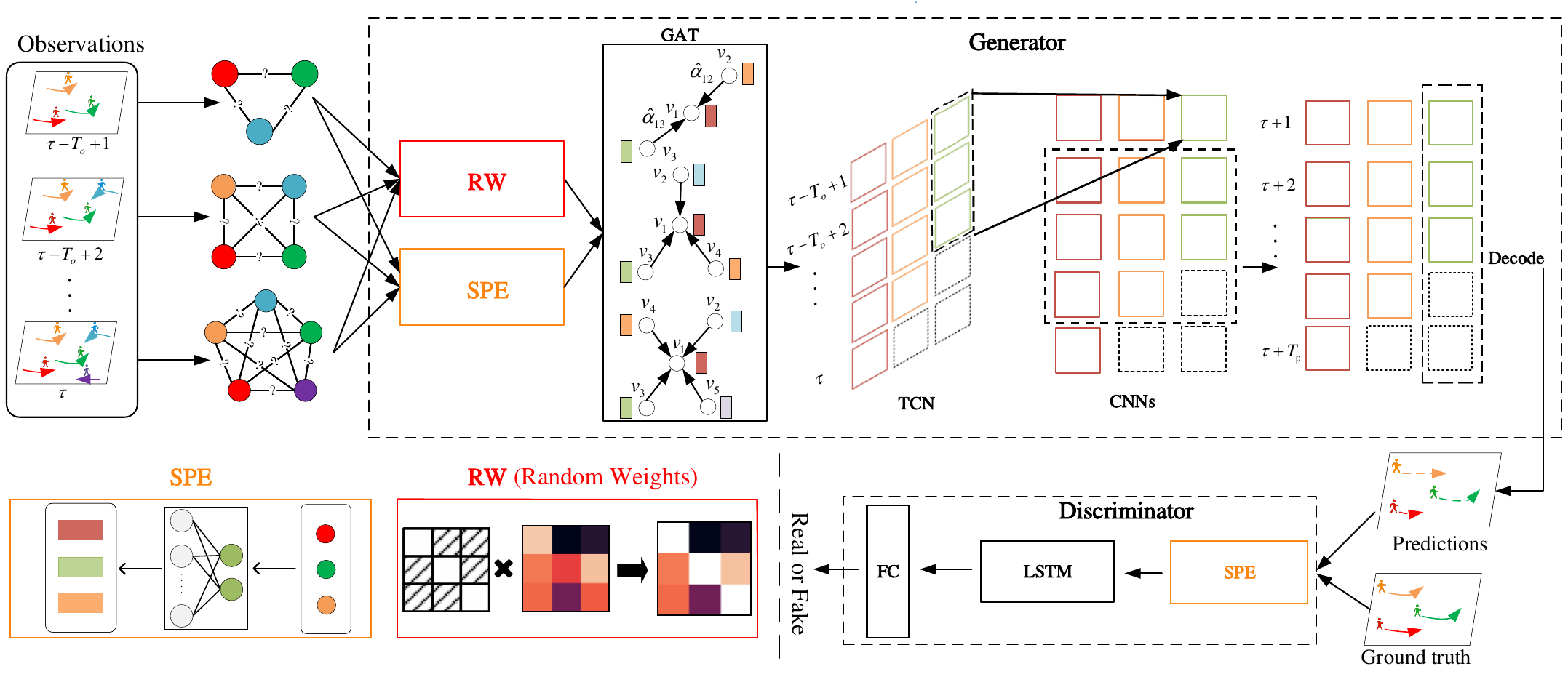}
    \caption{{\small The DTGAN framework consists of a Generator and a Discriminator. Spatial Embedding Layer (SPE) is used to embed node features, and Random Weights (RW) to randomly weight the adjacency matrix for each graph. 
    The Generator takes a set of graphs with node features and a random weights matrix as input.
    It utilizes the Graph Attention Network (GAT) to capture hidden node features and learn social interactions among pedestrians in the scene. Additionally, the Temporal Convolutional Network (TCN) with temporal dimension as the input channel extracts time sequence information, and the multi-layer Convolutional Neural Networks (CNNs) predict future trajectories.  Finally, a decoder is employed to obtain future trajectories. On the other hand, the Discriminator assesses both ground truth and predicted trajectories as input, classifying them as real or fake.}}
    \label{fig:model_pic}
\end{figure*}

\section{Methodology}
\label{section:method}
In this section, we present the design of the proposed DTGAN. As shown in Fig.~\ref{fig:model_pic}, the DTGAN comprises two main components: a Generator and a Discriminator. The Generator consists of a spatial embedding layer (SPE), a Graph Attention Network (GAT), Temporal Convolutional Networks (TCN), multi-layer Convolutional Neural Networks (CNNs), and a decoder. The Discriminator consists of an SPE, an LSTM network, and a Full Connection (FC).

\subsection{Problem Formulation}
Given a set of \szequation{$N$} pedestrians in a scene, the observed trajectory of pedestrian \szequation{$i$} over the past period \szequation{$T_o$} denotes as
\szequation{$ X_i = (x_i^t,y_i^t)$} from time-step \szequation{$t= \tau-To+1,\cdots,\tau$},
where \szequation{$(x_i^t,y_i^t)$} denotes the coordinate of pedestrian \szequation{$i$} at time \szequation{$t$}. The ground truth trajectory of pedestrian \szequation{$i$} over the future period \szequation{$T_p$} denotes as \szequation{$Y_i= (x_i^t,y_i^t)$} from time-step \szequation{$t = \tau+1,\cdots,\tau+T_p$} and the set of ground truth trajectories for all pedestrians denoted as \szequation{$\mathbf{Y}=\{ Y_1, Y_2,  \cdots, Y_N \}$}. The graph sequence data is constructed by observed pedestrian trajectory as shown in Fig. \ref{fig:graph_pic}, 
where each graph \szequation{$G^t = (V^t, A^t)$} consists of a set of pedestrians \szequation{$V^t $} and their social interactions represented by the adjacency matrix \szequation{$ A^t$}.
For any pair of pedestrians \szequation{$\forall i,j$} at time \szequation{$t$}, 
we define \szequation{$e_{ij}^t$} of \szequation{$A^t$} by \szequation{$e_{ij}^t = z \label{eq:randw} $} if and only if \szequation{$ i \neq j$}, where \szequation{$z$} is a random value. \szequation{$A^t$} is a randomly sampled weights matrix with zeros along the diagonal to account for the fact that social interactions occur between more than two individuals.
Our goal is that the Generator learns the data distribution with the help of adversarial training by taking the set of graphs \szequation{$\{ G^{\tau-T_o+1},\cdots, G^{\tau-1}, G^{\tau} \}$} as input, which predicts future trajectories of \szequation{$N$} pedestrians denoted as \szequation{$\hat{\mathbf{Y}}=\{\hat{Y}_1, \hat{Y}_2, \cdots, \hat{Y}_N \}$}.

\subsection{Generator}
Given a \szequation{$G^t$} at time \szequation{$t$}, we have \szequation{$v_i^t \in V^t$} and \szequation{$e_{ij}^t \in A^t$}. 
To capture feature interactions between nodes, we apply a spatial embedding layer (SPE) to obtain its embedding vector \szequation{$\vec{h}_i^t$}:

\szequation{
\begin{equation}
    \vec{h}{_i^t} = SPE \left(v_i^t,W_{emb}\right),\ v_i^t\in \mathbb{R}^D,\vec{h}{_i^t}\in \mathbb{R}^F,
    \label{eq:spe}
\end{equation}
}where \szequation{$SPE(\cdot)$} is an embedding using a Full Connection, \szequation{$W_{emb}$} is the embedding weight, \szequation{$D$} is the feature dimension of node \szequation{$v_i^t$}, \szequation{$F$} is the embedding dimension.

We introduce GAT \cite{velivckovic2018graph} to assign different weights to neighbor nodes when aggregating node information like previous methods\cite{huang2019stgat,kosaraju2019social,long2022unified}. 
The attention factor \szequation{$a_{ij}^t$} is calculated by feeding the embedding vector \szequation{$\vec{h}{_i^t}$} to the GAT:
\szequation{
\begin{equation}
    \alpha_{ij}^t = \frac{exp\left(LeakyReLU\left(\vec{a}^T\left[W\vec{h}{_i^t}||W\vec{h}{_j^t}\right]\right)\right)}
                         {\sum_{k\in N_i} exp\left(LeakyReLU\left(\vec{a}^T\left[W\vec{h}{_i^t}||W\vec{h}{_k^t}\right]\right)\right)},
\end{equation}
}where \szequation{$||$} denotes the splicing operation, \textit{LeakyReLU($\cdot$)} is a nonlinear activation function with negative-slope equal to 0.2, \szequation{$\vec{a}^T \in \mathbb{R}^{2F'}$} is a parameter of the feed-forward neural network. \szequation{$W$} is the linear transformation matrix. \par

To fully leverage the learning ability of the neural network, we redefine the attention factor \szequation{$\alpha_{ij}^t$} by introducing random weight \szequation{${A}^t$}:

\szequation{
\begin{equation}
    \hat{\alpha}_{ij}^t = \alpha_{ij}^t \cdot {e}_{ij}^t,
    \label{eq:9}
\end{equation}
}where \szequation{$\hat{\alpha}_{ij}^t$} is attention factor with a random weight. Then, we calculate a linear combination of their corresponding features and obtain the final output feature vector \szequation{$\vec{h'}{_i^{t}}$} of the node \szequation{$i$}:
\szequation{
\begin{equation}
    \vec{h'}{_i^{t}} = \sigma (\sum_{j \in N_i} \hat{\alpha}_{ij}^t W\vec{h}{_j^t}),
\end{equation}
}where \szequation{$\sigma(\cdot)$} is a non-linear activation function. The final output feature vectors set for all nodes is denoted as:

\szequation{
\begin{equation}
    \mathbf{H}^t = \left\{\vec{h'}{_1^{t}}, \vec{h'}{_2^{t}}, \cdots ,\vec{h'}{_N^{t}}\right\}.
\end{equation}
}

With the above definitions and steps, we obtain the final output feature vectors set for \szequation{$\{ G^{\tau-T_o+1},\cdots, G^{\tau-1}, G^{\tau} \}$}, denoted as:
\szequation{
\begin{equation}
    \mathbf{H} = \left\{H^{\tau-T_o+1}, \cdots, H^{\tau-1}, H^{\tau}\right\}  \in  \mathbb{R}^{T_0 \times N \times F},
\end{equation}
}where \szequation{$\mathbf{H}$} is the set of output feature vectors in GAT.

The Temporal Convolutional Network (TCN) is regarded as a natural starting point and a powerful toolkit for sequence modeling \cite{TCN}. We utilize \szequation{$TCN(\cdot)$} to extract the time sequence information by inputting \szequation{$\mathbf{H}$}. Subsequently, we employ Multilayer Convolutional Neural Networks \szequation{$CNNs(\cdot)$} for prediction, with an input channel of size \szequation{$T_o$} and an output channel of size \szequation{$T_p$}.

\szequation{
\begin{equation}
    \mathbf{V} = CNNs (TCN (\mathbf{H},W_{tcn}), W_{cnns}),
\end{equation}
}where \szequation{$\mathbf{V} \in \mathbb{R}^{T_p \times N \times F}$}
is the set of future node features predicted, \szequation{$W_{tcn}$} and \szequation{$W_{cnns}$} are the parameters of the network.

Finally, we decode \szequation{$\mathbf{V}$} using a CNN to obtain the future trajectories of all pedestrians:
\szequation{
\begin{equation}
\begin{array}{rcl}
    \begin{aligned}
        \hat{\mathbf{Y}} & = & Decoder(\mathbf{V},\ W_{dec}) ,
    \end{aligned}
\end{array}
\end{equation}
}where \szequation{$\hat{\mathbf{Y}} \in \mathbb{R}^{T_p \times N \times D}$} is predicted trajectories of $N$ pedestrians over the future period \szequation{$T_p$}.

\begin{table}[htbp]
    \caption{Notation description}
    \centering
    \begin{tabular}{cl}
        \toprule	
        \textbf{\emph{Symbols}} &  \textbf{\emph{Meaning}} \\
        \toprule
        $N$ & \textit{number of pedestrians in a scene}\\
        $i$ & \textit{the pedestrian $i$}\\
        $T_o$ & \textit{period of observed trajectory}\\
        $T_p$ & \textit{period of future trajectory}\\
        $X_i$ & \textit{observed trajectory of pedestrian $i$}\\
        $Y_i$ & \textit{ground truth future trajectory of pedestrian $i$}\\
        $\hat{Y}_i$ & \textit{predicted future trajectory of pedestrian $i$}\\
        $V^t$ & \textit{graph nodes which represents a set of pedestrians at time $t$}\\
        $A^t$ & \textit{adjacency matrix which represents social interactions} \\ 
              & \textit{between pedestrians at time $t$}\\
        $G^t$ & \textit{graph constructed from 2D pedestrian coordinates} \\
              & \textit{at time $t$, equal to $(V^t,A^t)$}\\
        $(x_i^t,y_i^t)$ & \textit{coordinate of the pedestrian $i$ at time $t$}\\
        $\tau$ & \textit{the current moment of a trajectory}\\
        $\hat{\alpha}_{ij}$ & \textit{attention coefficient with a random weight}\\
    
    \bottomrule
    \label{tab:notation}
    \end{tabular}
\end{table}

\subsection{Discriminator}
As a Discriminator, we select a prove-optimized configuration as the one at the cores of \cite{ace} and \cite{romero2023dragon}, which matches the encode-decode characteristics of the Generator.
It mainly consists of a spatial embedding layer, an LSTM network, and a Full Connection. 
To maintain consistency with the Generator, we employ an SPE to embed the trajectories, mapping them to a high-dimensional feature representation. Next, we input this representation into an LSTM network to obtain the last hidden state. Finally, the last hidden state is passed through a Fully Connected Network with the ReLU activation function to score both the ground truth and predicted trajectories.


\subsection{Loss Function}
GAN is a generative model that consists of a Generator and a Discriminator. The Generator is trained to produce realistic samples, while the Discriminator is trained to distinguish between ground truth samples and generated samples. Particularly, the Discriminator prefers to encourage the Generator to produce samples that are diverse and realistic rather than samples maintaining alignment to ground truth.
To balance both realism and accuracy in predictions, we combine task loss and adversarial loss during adversarial training. This offers feedback to the Generator on the realism of predictions while also measuring the difference between predictions and ground truth.
By doing so, the Generator can learn to produce samples that not only resemble the ground truth but also maintain the diversity and realism of the underlying distribution.
\par

\textbf{DTGAN} To avoid model collapse of standard GAN~\cite{goodfellow2014generative}, we leverage the WGAN~\cite{arjovsky2017wasserstein} training mechanism for DTGAN, which has proved no variant of GAN consistently outperforms the standard GAN by Huang et al.~\cite{NEURIPS2018_GAN_equal}.
The loss function of WGAN is denoted as:
\szequation{
\begin{equation}
    \mathcal{L}_{DTGAN} = \mathbb{E}_{\hat{Y}\sim p_g}[D(\mathbf{\hat{Y}})] - \mathbb{E}_{Y \sim p_d}[D(\mathbf{Y})],
\label{eq:DTGAN}
\end{equation}
}where \szequation{$D(\hat{\mathbf{Y}})$} is the discriminant result of the predicted trajectories \szequation{$\hat{\mathbf{Y}}$}, $p_g$ is the distribution of the predicted trajectories, \szequation{$D(\mathbf{Y})$} is the discriminant result of the ground truth trajectories \szequation{$\mathbf{Y}$}, and $p_d$ is the distribution of the ground truth trajectories. \par

\textbf{DTGAN-M} Mean Square Error (MSE) is the most commonly used error in the regression loss function, so we introduce MSE in Equation (\ref{eq:DTGAN}) as the loss function of DTGAN-M. Motivated by \cite{gupta2018social},  we predict \szequation{$K$} future trajectories and use the trajectory with the smallest trajectory displacement error as our prediction. We calculate MSE between the ground truth trajectories and the predicted trajectories. The loss function of DTGAN-M is denoted as:
\szequation{
\begin{equation}
\begin{split}
    &\mathcal{L}_{M} = \min_K ||\hat{Y}_i^k-Y_i||_2,  \\
    &\mathcal{L}_{DTGAN-M}=\mathop{\min}_G \mathop{\max}_D  \mathcal{L}_{DTGAN} + \gamma \mathcal{L}_{M},
\end{split}
\end{equation}
}where $\hat{Y}_i^k$ is $k_{th}$ predicted trajectories of pedestrian $i$, $Y_i$ is the ground truth trajectory of pedestrian $i$, $\gamma$ is the hyper-parameters.  \par

\textbf{DTGAN-G}  Maximum Likelihood Estimation (MLE) achieves significant results in previous works \cite{alahi2016social,mohamed2020social}, which assumes that trajectory coordinates $(x_i^t,y_i^t)$ follow bi-variate Gaussian distribution.  Naturally, we introduce the negative log-likelihood function in Equation (\ref{eq:DTGAN}) as the loss function of DTGAN-G.
The loss function of DTGAN-G is denoted as:

\szequation{
\begin{equation}
\begin{split}
	\mathcal{L}_{G} =\mathop{\arg\min}_{\mu,\sigma,\rho} \Big(-\sum_{t=1}^{T_p}log\big(P\left((x_i^t,y_i^t)|\mu_i^t,\sigma_i^t,\rho_i^t\right)\big)\Big), \\
    \mathcal{L}_{DTGAN-G} = \mathop{\min}_G \mathop{\max}_D \mathcal{L} _{DTGAN} + \gamma \mathcal{L}_{G},
\end{split}
\end{equation}
}where $\mu_i^t$ is  the mean of the distribution, $\sigma_i^t$ is the variances and $\rho_i^t$ is the correlation coefficient.  \par

\textbf{DTGAN-U} Considering the walking range of pedestrians is limited by their speeds. We assume that trajectory coordinates $(x_i^t,y_i^t)$ may follow bi-variate Uniform distribution. Specifically, the coordinates of the pedestrian at the next time may be distributed in the area of a circle with radius $\hat{r}$. Consequently, the probability density function with respect to coordinate $(x_i^t,y_i^t)$ is denoted as:

\szequation{
\begin{equation}
    p(x_i^t,y_i^t|U(\hat{r}))=\left\{
    \begin{aligned}
\frac{1}{\pi \hat{r}^2+\epsilon}, & \quad if \ \sqrt{{x_i^t}^2+{y_i^t}^2} < \hat{r} \\
0, & \quad else
\end{aligned}
\right. ,
\label{eq:p}
\end{equation}
}where $U(\hat{r})$ is the predicted Uniform distribution of a circle area with the radius is $\hat{r}$, $\epsilon$ is added to avoid dividing by zero.
We minimize the negative log-likelihood function of Equation (\ref{eq:p}) and introduce it in Equation (\ref{eq:DTGAN}) as the loss function of DTGAN-U:

\szequation{
\begin{equation}
\begin{split}
    \mathcal{L}_{U} = argmin(-\prod_{t=1}^{T_p}log(P(x_i^t,y_i^t)), \\
    \mathcal{L}_{DTGAN-U} = \mathop{\min}_G \mathop{\max}_D \mathcal{L} _{DTGAN} + \gamma \mathcal{L}_{U}.
\end{split}
\end{equation}
}


\begin{table*}[t]
    \renewcommand\arraystretch{0.9}
    \centering
    \caption{{\small  The performance of different methods in terms of ADE / FDE and AMD/AMV metrics. AVG is the average results for each model over all datasets.  M is a non-reported model. The \textbf{bold} shows the best of AMD/AMV, and the \underline{underline} shows the best of ADE/FDE. For all metrics, the lower the better.}}
    \resizebox{0.9\linewidth}{!}{
    \begin{tabular}{ccc|ccccc|c}
        \toprule[1pt]

         Baselines & Year & Metric & ETH & HOTEL & UNIV & ZARA1 & ZARA2 & AVG \\
        \toprule[1pt]
                \toprule[1pt]
         Social-LSTM\cite{alahi2016social}  & 2016 & \makecell{ADE/FDE \\ AMD/AMV}  & \makecell{1.09/2.35 \\M/M } 
        & \makecell{0.79/1.76\\M/M}  & \makecell{0.67/1.40\\M/M}    & \makecell{0.47/1.00\\M/M}    & \makecell{0.56/1.17\\ M/M}  & \makecell{0.72/1.54 \\M}      \\ 
        \hline               
        S-GAN\cite{gupta2018social} & 2018 & \makecell{ADE/FDE \\ AMD/AMV}  & \makecell{0.81/1.52 \\ 3.94/0.373 } 
        & \makecell{0.72/1.61\\2.59/0.384}  & \makecell{0.60/1.26\\2.37/0.440}    & \makecell{0.34/0.69\\1.79/0.355}    & \makecell{0.42/0.84\\1.66/0.254}  & \makecell{0.58/1.18\\1.42}      \\ 
        \hline       
         social-BiGAT\cite{kosaraju2019social} & 2019 & \makecell{ADE/FDE \\ AMD/AMV}  & \makecell{0.69/1.29\\M/M}  & \makecell{0.49/1.01\\M/M}    & \makecell{0.55/1.32\\M/M}    & \makecell{0.30/0.62\\M/M}  & \makecell{0.36/0.75\\M/M}    
       & \makecell{0.48/1.00 \\ M }  \\ 
        \hline
      SoPhie\cite{sadeghian2019SoPhie}  & 2019 & \makecell{ADE/FDE \\ AMD/AMV}  & \makecell{0.70/1.43 \\ M/M}  & \makecell{0.76/1.67\\M/M}  & \makecell{0.54/1.24\\M/M}    & \makecell{0.30/0.63\\M/M}    & \makecell{0.38/0.78\\M/M}  & \makecell{0.54/1.15\\M}      \\ 
        \hline
        STGAT\cite{huang2019stgat}  & 2019 & \makecell{ADE/FDE \\ AMD/AMV}  & \makecell{0.68/1.29\\ M/M}      &\makecell{ 0.68/1.40\\ M/M}      &\makecell{ 0.57/1.29\\ M/M}      &\makecell{ 0.29/0.60 \\ M/M }  &\makecell{ 0.37/0.75 \\ M/M }  &\makecell{ 0.52/1.07 \\ M }\\ \hline
        
        Social-STGCNN\cite{mohamed2020social} & 2020 & \makecell{ADE/FDE \\ AMD/AMV}  & \makecell{0.64/1.11 \\ 3.73/0.09} & \makecell{ 0.49/0.85\\\textbf{1.67}/0.30}  &\makecell{ 0.44/0.79\\3.31/0.06}  &\makecell{ 0.34/0.53\\ \textbf{1.65}/0.15}   &\makecell{ 0.30/0.48\\1.57/0.10}   &\makecell{ 0.44/0.75\\1.26}   \\ 
        \hline

        STAR\cite{yu2020spatio}  & 2020 & \makecell{ADE/FDE \\ AMD/AMV}  & \makecell{0.56/1.11\\ M/M}      &\makecell{ 0.26/0.50\\ M/M}      &\makecell{ 0.52/1.15\\ M/M}      &\makecell{ 0.41/0.90 \\ M/M }  &\makecell{ 0.31/0.71 \\ M/M }  &\makecell{ 0.41/0.87 \\ M }\\ \hline

        TPNMS\cite{liang2021temporal}    & 2021 & \makecell{ADE/FDE \\ AMD/AMV}  & \makecell{\underline{0.52}/\underline{0.89}\\ 28.04/1.26 }      &\makecell{ \underline{0.22}/ \underline{0.39}\\ 23.25/0.29}      &\makecell{ 0.55/1.13\\ 64.70/0.11}      &\makecell{ 0.35/0.70 \\ 23.49/0.26}  &\makecell{ 0.27/0.56 \\ 49.99/0.13}  &\makecell{ 0.38/0.73 \\ 19.15 }\\ \hline

                
        BR-GAN\cite{pang2022br}  & 2022 & \makecell{ADE/FDE \\ AMD/AMV}  & \makecell{0.73/1.37 \\ M/M} & \makecell{0.55/1.13\\M/M}  &\makecell{0.53/1.07\\ M/M}  &\makecell{0.35/0.71 \\ M/M}   &\makecell{ 0.35/0.72\\ M/M}   &\makecell{ 0.50/1.00\\ M}   \\ 
        \hline

        SocialDualCVAE\cite{gao2022social}  & 2022 & \makecell{ADE/FDE \\ AMD/AMV}  & \makecell{0.66/1.18 \\ M/M} & \makecell{0.34/0.61\\M/M }  &\makecell{0.39/0.74\\M/M }  &\makecell{\underline{0.27}/0.48 \\ M/M}   &\makecell{ \underline{0.24}/0.42\\ M/M}   &\makecell{ 0.38/0.69\\ M}   \\ 
        \hline


    \textbf{ DTGAN (Ours)}  & - & \makecell{ADE/FDE \\ AMD/AMV} & \makecell{0.68/1.43\\ 10.35/\textbf{0.04} } & \makecell{0.30/0.52 \\ 6.10/0.12 } & \makecell{ 0.51/1.07\\ 20.18/\textbf{0.003}} &\makecell{ 0.31/0.67\\6.86/ \textbf{0.02}} & \makecell{ 0.28/0.59\\5.69/\textbf{0.02}} 
        & \makecell{0.42/0.85\\ 4.93 } \\
         \hline

    \textbf{ DTGAN-G (Ours)} & - & \makecell{ADE/FDE \\ AMD/AMV} & \makecell{0.65/1.22\\ \textbf{2.82}/0.14 } & \makecell{0.29/ 0.40 \\ 1.78/\textbf{0.12} } & \makecell{ \underline{0.33}/ \underline{0.54}\\ \textbf{1.10}/0.47 } &\makecell{ 0.30/ \underline{0.48}\\1.74/0.09} & \makecell{ 0.26/ \underline{0.42}\\ \textbf{1.51}/0.07} 
         & \makecell{\underline{0.36}/\underline{0.61}\\ \textbf{0.98} } \\
        
        \hline
    \end{tabular}
    }

    \label{tab:baseline}
\end{table*}

\begin{figure*}[ht]
    \centering
    \includegraphics[width=7.0in]{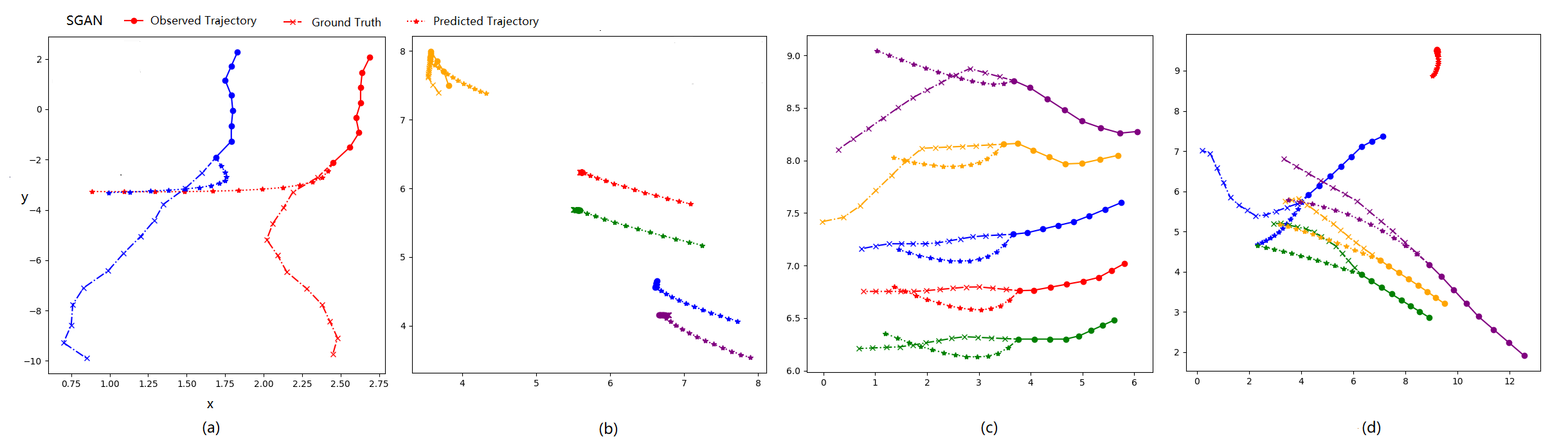}

     \includegraphics[width=7.0in]{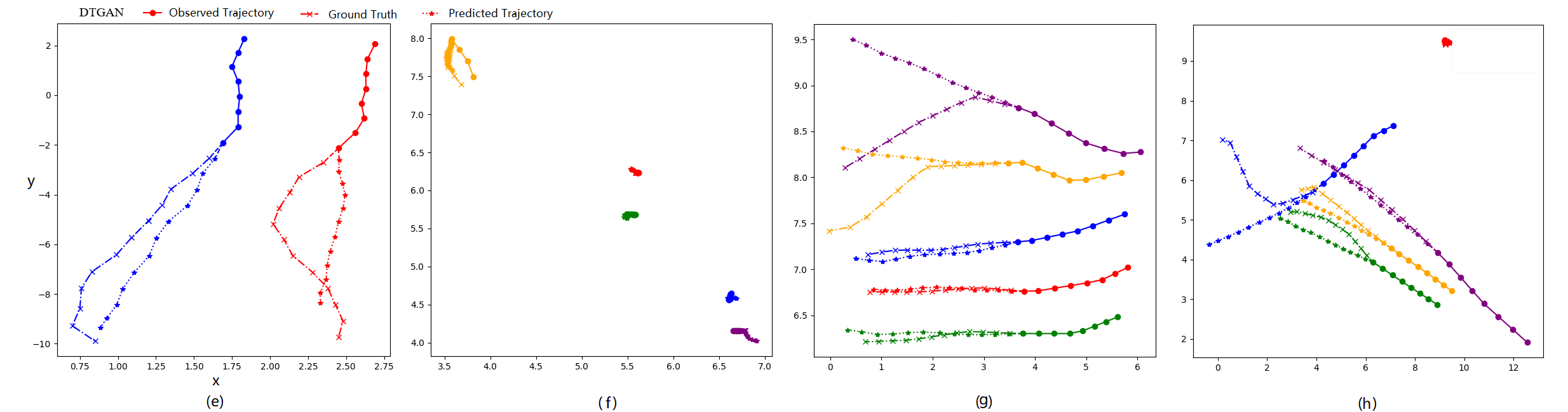}

    \caption{{\small Illustration of single trajectory prediction. We use the coordinate system to represent the plane position of trajectory points, y is the vertical axis coordinate, x is the horizontal axis coordinate, and the trajectory point represents each moment. Two models use the best amongst 20 samples for evaluation. Note that the coordinate origin in each subplot is not exactly the same, and the intersection point does not necessarily collide.}}
    \label{fig:traj_one}
\end{figure*}

\section{Experiments}
\label{section:Experiments}
Our framework is evaluated on two publicly available pedestrian trajectory datasets: The ETH \cite{pellegrini2009you} (including ETH and HOTEL) and UCY \cite{leal2014learning} (including UNIV, ZARA1, and ZARA2).
The ETH dataset contains more straight trajectories with few social interactions, and the UCY dataset contains multi-versatile trajectories in crowded scenes with more social interactions.
All datasets are a total of 2,206 pedestrian trajectories and 4 different scenes, with a sampling interval of 0.4 seconds. There are complex scenes such as following others, dispersing from a specific point, merging from different directions, avoiding collision, and walking in groups.

\subsection{Evaluation Metrics}
We evaluate DTGAN on two generic metrics: Average Displacement Error (ADE) and Final Displacement Error (FDE), which are reported in previous works \cite{alahi2016social,mohamed2020social}.

Furthermore, we introduce another two metrics: Average Mahalanobis Distance (AMD) and Average Maximum Eigenvalue (AMV), which have proven more suitable for generated models \cite{mohamed2022social}.\par 
AMD measures the distance from a point to a distribution and also correlates the distance with the predicted variance.
\szequation{
\begin{equation}
    \text{AMD} = \frac{1}{N \times T_p} \sum_{n \in N} \sum_{t \in T_p} M_D(\hat{\mu}^{GMM,t}_i, \hat{G}_i^t, (x_i^t,y_i^t)),
\end{equation}
}where $\hat{\mu}$ is the mean of the GMM distribution, $\hat{G}$ is the variance. AMV evaluates the overall spread of the predicted trajectories:
\szequation{
\begin{equation}
     \text{AMV} = \frac{1}{N \times T_p} \sum_{n \in N} \sum_{t \in T_p} \lambda_1^\downarrow(\hat{\Sigma}^{\text{GMM},t}_i),
\end{equation}
}where $\lambda_1^\downarrow$ is the eigenvalue of the matrix with the largest amplitude, $\hat{\Sigma}_{GMM}$ is the covariance matrix of the predicted GMM distribution.
\subsection{Implementation Details}\label{sec:EB}
We use the same data processing method as \cite{mohamed2020social}, and we divide the model training into two parts: pre-training \cite{lei2019gcn-gan}   and adversarial training \cite{arjovsky2017wasserstein}.
Since we use random weights, pre-training allows the Generator to learn potential interaction information about pedestrians in advance. The batch size is set to 32. 
For pre-training, the learning rate is set to 0.001. For adversarial training, the learning rate of the Generator and the Discriminator are both set to 0.00001, the Generator gradient is truncated to [-1.0,1.0], and the Discriminator weight is truncated to [-0.1,0.1]. The size of the convolution kernel for the spatial decoder is set to 3. 
In all our experiments, we predict the next 12 frames using 8 observed frames and utilize the leave-one-out cross-validation strategy \cite{alahi2016social} to train and test.
Errors reported are ADE/FDE metrics and AMD/AMV metrics.
ADE and FDE take the smallest value out of 20 samples.
AMD and AMV take the smallest value out of 100 samples. The experiments
are conducted based on the MindSpore framework platform.

\subsection{Baselines}
We compare against the following baselines:
Social-LSTM\cite{alahi2016social}, 
Social-STGCNN\cite{mohamed2020social}, SocialDualCVAE \cite{gao2022social}, 
S-GAN\cite{gupta2018social}, 
Social-BiGAT\cite{kosaraju2019social}, SoPhie\cite{sadeghian2019SoPhie}, TPNMS\cite{liang2021temporal},
BR-GAN\cite{pang2022br}, 
STGAT\cite{huang2019stgat}, 
STAR\cite{yu2020spatio}.
For consistency with previous works, DTGAN generates 20 samples from the Generator, and DTGAN-G extracts 20 samples from predicted distributions.

\begin{figure*}[ht]
    \centering
    \includegraphics[width=7.0in]{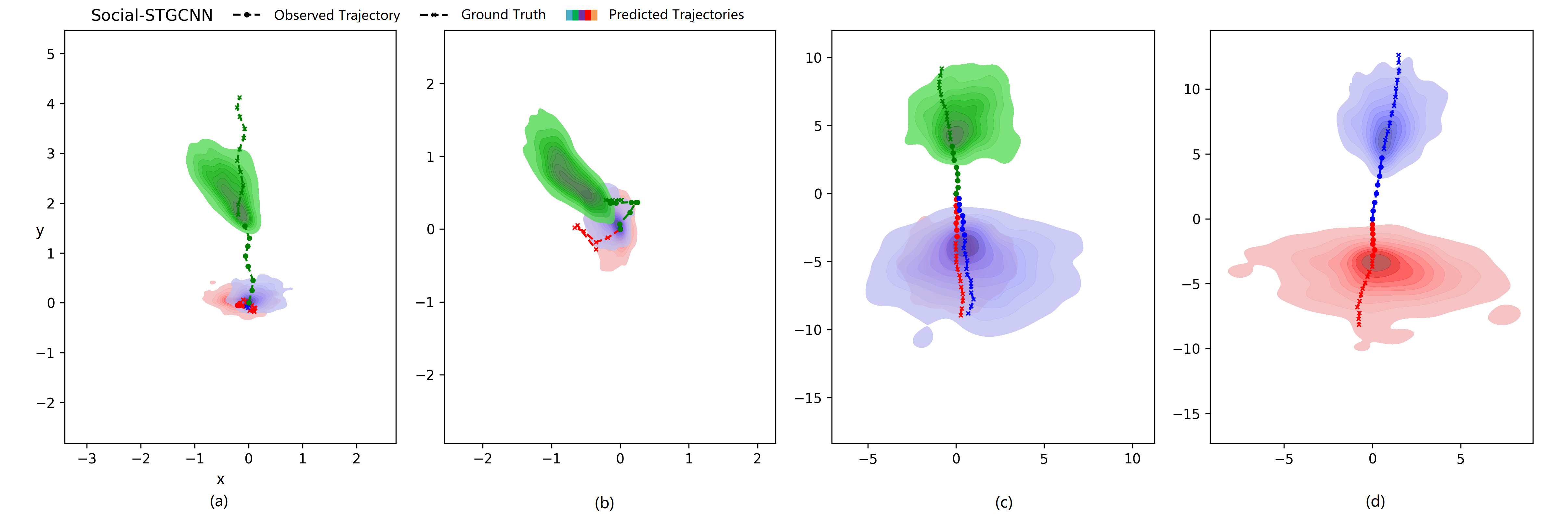}

    \includegraphics[width=7.0in]{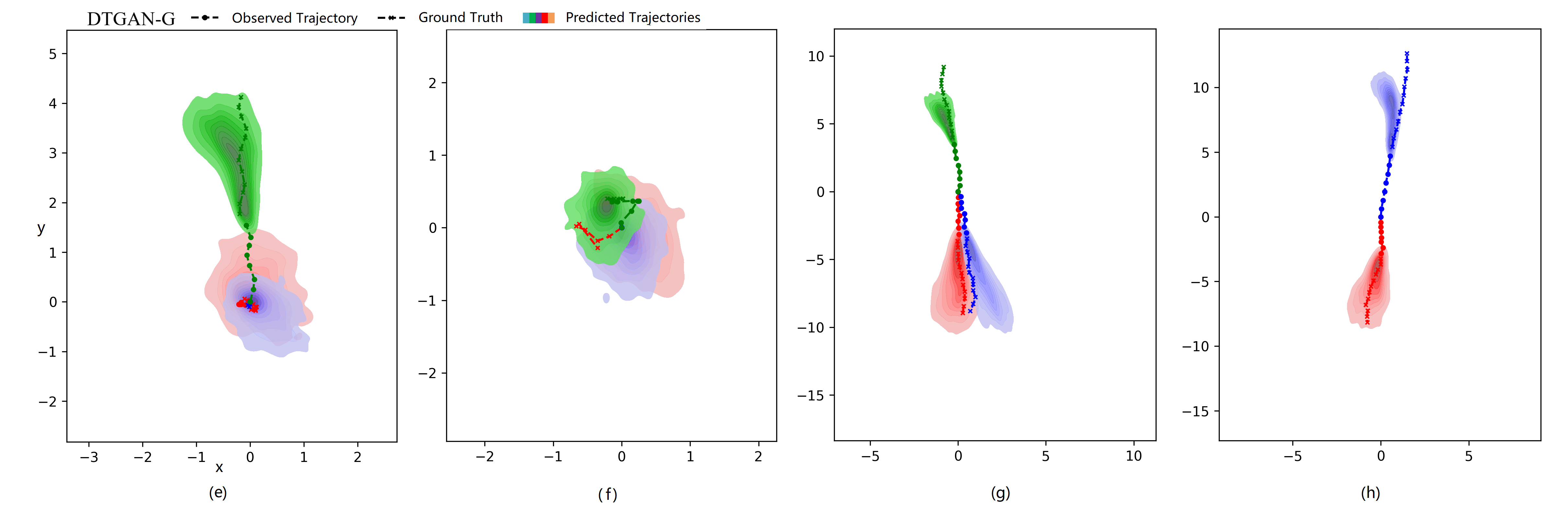}
    \caption{{\small Illustration of trajectory distribution prediction. Each pedestrian is assigned various colors. The colored area of the ellipse represents the probability density distribution from the prediction. The wider the area of the ellipse, the greater the variance.}}
    \label{fig:traj_dist}
\end{figure*}
\subsection{Quantitative Analysis}
We report the ADE/FDE and AMD/AMV metrics on different baselines using the ETH/UCY datasets as shown in Table \ref{tab:baseline}.
In contrast to approaches like social-BiGAT, SoPhie, and STGAT, which employ GAT directly on hidden states, DTGAN incorporates GAT into graphs to learn pedestrian interactions. This strategy yields notable improvements in performance across all datasets in terms of ADE/FDE.
This suggests that utilizing node relationships for modeling interactions enhances the neural network's understanding of pedestrian dynamics.
Furthermore, when comparing the modeling of pedestrian interactions using node relationships in the graph to the Social-STGCNN model, DTGAN demonstrates comparable ADE/FDE performance to Social-STGCNN across all datasets.
Notably, DTGAN even outperforms Social-STGCNN in terms of average ADE.
DTGAN-G, which combines the log-likelihood function in adversarial training, yields a 16.7$\%$ improvement in ADE and a 39.3$\%$ improvement in FDE.  Moreover, it achieves superior average performance in terms of AMD/AMV compared to all baselines. This suggests that the incorporation of task loss functions during adversarial training is a valuable and effective exploratory approach.
Notably, DTGAN-G performs better on UCY datasets than on ETH datasets, implying its effectiveness in capturing social interactions to improve prediction accuracy, particularly in scenes with more complex interactions.
While TPNMS exhibits similar ADE/FDE values to DTGAN-G, there is a significant difference in the quality of generated samples, as indicated by AMD/AMV metrics. 
we observe that TPNMS generates a tight distribution that does not surround the ground truth. 
Furthermore, on UCY datasets with more interactions, DTGAN and DTGAN-G outperform all GAN-based baselines in terms of ADE/FDE. This finding further underscores that neural networks more readily comprehend modeling interaction among pedestrians by using graph structure data.
In summary, DTGAN-G demonstrates the best performance among all baselines in terms of average ADE/FDE and AMD/AMV.
\par
\subsection{Qualitative Analysis}
In this section, we provide a qualitative comparison with classic models S-GAN and Social-STGCNN to demonstrate how our framework effectively captures pedestrian social interactions for accurate predictions.

\subsubsection{DTGAN \textit{vs.} S-GAN} S-GAN is the most classic of the generative models in this field, we compare them from four dimensions: walking direction, staying, companionship, and collision avoidance.

\noindent \textbf{\textit{Walking direction}}: As shown in Fig. \ref{fig:traj_one} (a)(e), DTGAN accurately predicts the pedestrian walking direction and final location coordinates, while S-GAN predicts direction deviates significantly from the true direction, gradually moving away from the destination. \par
\noindent \textbf{\textit{Staying}}: When pedestrians stop or move in a small area, Fig. \ref{fig:traj_one} (b)(f) shows that DTGAN gives reasonable predictions. Conversely, S-GAN often misinterprets such scenarios as walking motions. In addition, DTGAN also learns the complexities of pedestrians turning around and returning, as indicated by the color {\#FFA500} trajectory. \par
\noindent \textbf{\textit{Companionship}}: In Fig. \ref{fig:traj_one} (c)(g), we observe that five individuals are closely clustered, within a distance of approximately 0.5 meters. The color {\#7F007F} and color {\#FFA500} trajectories are two individuals' companionship and color {\#0000FF}, color {\#FF0000} and color {\#007A00} trajectories are three individuals' companionship.
The results show that S-GAN wrongly identifies the five closely spaced people as companions, resulting in similar predictions.
On the other hand, DTGAN successfully distinguishes between two different groups. Despite this, both models' predicted paths deviate from the ground truth. We conjecture 
that the sudden changes in pedestrian trajectories could potentially be attributed to external factors, such as obstacles, that are not considered within the models. \par
\noindent \textbf{\textit{Collision Avoidance}}: As we know, it is human nature to avoid collisions, but in Fig. \ref{fig:traj_one} (d)(h), a collision occurs between S-GAN's color {\#0000FF} and color {\#007A00} trajectories at time-step $t = 12$. We notice such a phenomenon is not rare in S-GAN. In contrast, DTGAN consistently offers collision-free path predictions. \par

\subsubsection{DTGAN-G \textit{vs.} Social-STGCNN}
Social-STGCNN also employs graph nodes to simulate pedestrian interactions but uses pre-defined weight. We compare them in terms of uncertainty trajectories of stopped pedestrians and certainty trajectories of walking pedestrians.  \par
\noindent \textbf{\textit{Uncertainty trajectories of stopped pedestrians}}:
When pedestrians wander or stop in place, their uncertainty about future trajectory is higher, which means distribution variance is greater. Fig. \ref{fig:traj_dist} (a)(e) shows that DTGAN-G
has greater distribution variance than Social-STGCNN involving stopped pedestrians as shown in color {\#0000FF} and color {\#FF0000}.  Its predicted density distribution closely matches the ground truth trajectory involving walking pedestrians as shown in color {\#007A00}.
On the other hand, in Fig. \ref{fig:traj_dist} (b)(f),
we find that Social-STGCNN tends to predict more deterministic trajectories of pedestrians when they wander or stay in place, while it should be wandering or staying more likely.
\par
\noindent \textbf{\textit{Certainty trajectories of walking pedestrians}}: Fig. \ref{fig:traj_dist} (c)(g) shows pedestrians walking in opposite directions, DTGANG exhibits more accurate directionality compared to Social-STGCNN. Furthermore, in the ellipse area of Fig. \ref{fig:traj_dist} (c)(g)(d)(h), the variance of DTGAN-G is smaller, which means that it has higher confidence in predicting the future trajectory of pedestrians. \par
\sethlcolor{yellow}Based on the qualitative analysis performed, our results indicate that DTGAN-G exhibits a superior ability to understand the intentions of pedestrians compared to Social-STGCNN. 
\par

\begin{figure}[ht]
    \centering
    \includegraphics[scale=0.4]{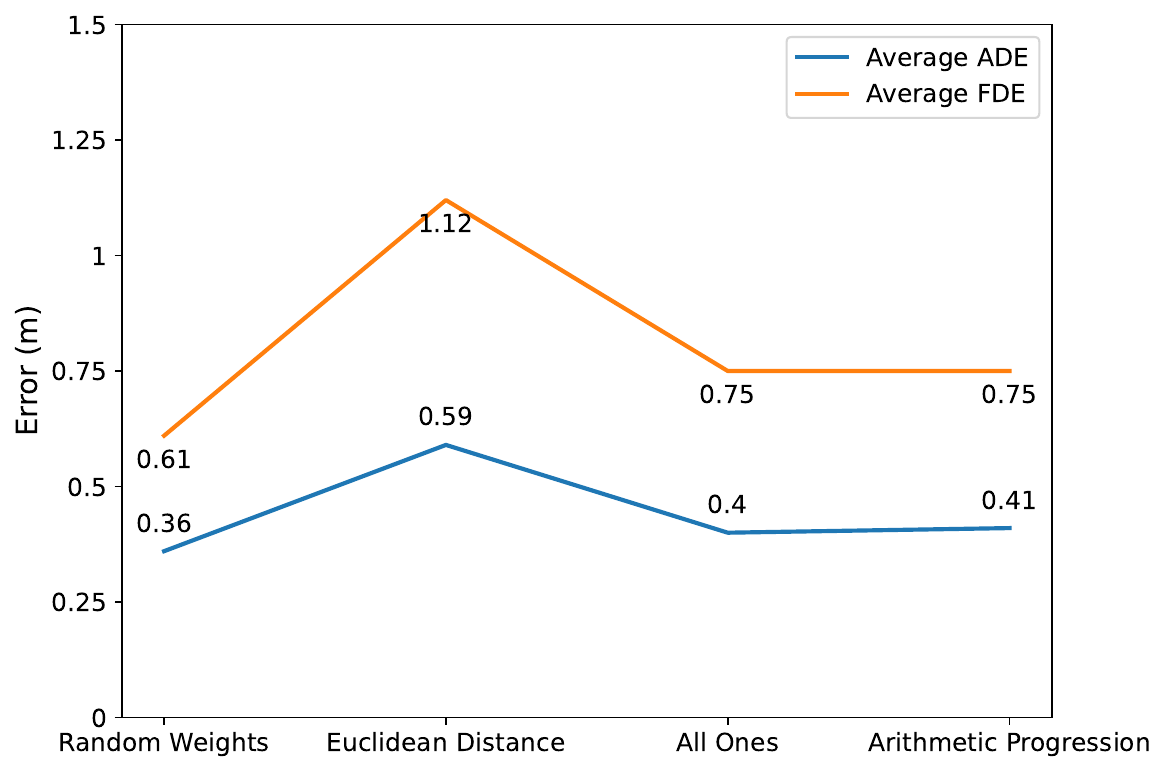}
    \caption{\small Average ADE/FDE of all datasets for different generated ways of weights.
 }
    \label{fig:different weights}
\end{figure}

\begin{table}[h]
    \centering
    \caption{\small ADE/FDE results for using different task loss functions in GAN. }
    \begin{tabular}{c|c|c|c|c}
        \toprule[1pt]
         & DTGAN & DTGAN-M  & DTGAN-G & DTGAN-U \\
        \midrule
        \midrule
        ETH     & 0.68/1.43 & 0.79/1.60 & 0.65/1.22 & 2.84/4.81\\
        HOTEL   & 0.30/0.52 & 0.33/0.65 & 0.29/0.40 & 1.15/2.09 \\        
        UNIV	&0.52/1.07	&0.55/1.06	&0.33/0.54 &1.22/2.22 \\
        ZARA1	&0.31/0.67	&0.35/0.76	&0.30/0.48 &2.50/4.61 \\
        ZARA2	&0.28/0.59	&0.29/0.60	&0.26/0.42 &1.38/2.53 \\
        \midrule
        AVG		&0.42/0.85	&0.46/0.94	&\textbf{0.36}/\textbf{0.61} &1.82/3.25 \\
        \hline

    \end{tabular}
    
    \label{tab:loss functions}
\end{table}


\subsection{Robustness Analysis about Random Weights}
We pre-define weights to validate the performance of DTGAN-G, as shown in Fig. \ref{fig:different weights}. 
Random Weights is a matrix with the value randomly generated by the computer. Euclidean Distance is a matrix with the value of the reciprocal of the Euclidean distance between trajectory points. All Ones matrix is a matrix with all values of one. Arithmetic Progression is a matrix obtained by adding a fixed constant to the preceding term.
From the result, Random Weights is the best performance, but 
Euclidean Distance has the highest average ADE and FDE. 
This is attributed to the Euclidean Distance approach treating pedestrians equidistant in front and behind as equally significant. However, human behavior gives more importance to pedestrians in the forward direction during walking. This reflects the limitation of pre-defined rules.

To further analyze the sensitivity for randomness changes, we generate five different sets of random weights using different random seeds to train and evaluate our model on each of the five subsets. We then calculate the average ADE and FDE on each subset across the five random weights sets of experiments to obtain the final results reported in Table \ref{tab:random}. It is evident that the mean values of ADE and FDE across 5 random seeds are consistent with the results in Table \ref{tab:baseline}, and the standard variance is slight.  The maximum fluctuation range of average ADE/FDE is only 7 cm in terms of standard variance. This indicates that our framework maintains stability even when there are changes in randomness.

\begin{figure}
    \centering    
    \includegraphics[scale=0.45]{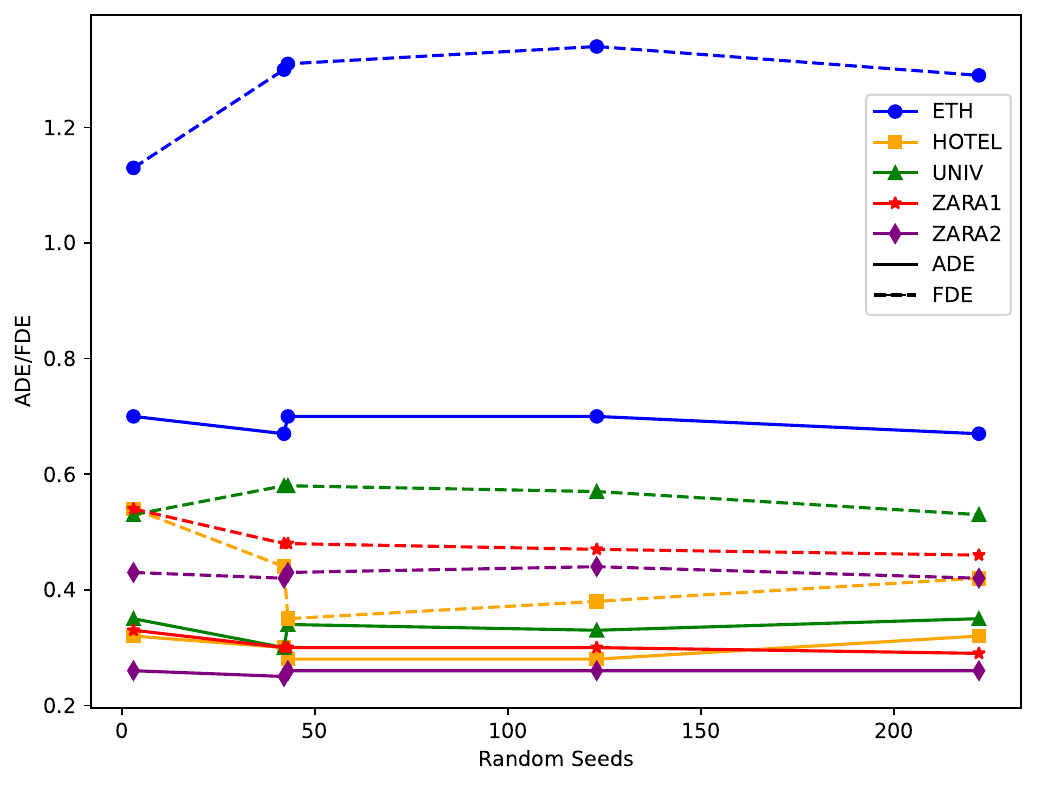}
    \caption{Line chart of ADE/FDE using various random seeds. The x-axis represents the number of random seeds, while the y-axis represents the ADE/FDE values. Different line types and colors are used to represent different datasets. Solid and dashed lines indicate ADE and FDE, respectively.}
    \label{fig:line}
\end{figure}

The experimental results about the robustness of random weights are visualized in Fig \ref{fig:line}. 
When the random seed is set to 42 or 43, respectively, the data points appear largely overlap.  While there may be slight fluctuations observed in the lines, it is not indicative of instability in the model. Overall, the consistent trend and minimal variations in the ADE/FDE values suggest that the model performs reasonably well across different random seed settings. 
Each density curve on the plot represents the distribution of values for each dataset.
\begin{table}[h]
    \centering
    \caption{ADE/FDE results for using different random seeds.}
    \setlength{\tabcolsep}{0.5mm}{
    \begin{tabular}{c|c|c|c|c|c}
        \toprule
          & ETH & HOTEL & UNIV & ZARA1 & ZARA2 \\
         \midrule
         \specialrule{0em}{1pt}{1pt}
         seed(3) & 0.70/1.13 & 0.32/0.54 & 0.35/0.53 & 0.33/0.54 & 0.26/0.43 \\
         
         seed(42) & 0.67/1.30 & 0.30/0.44 & 0.30/0.58 & 0.30/0.48 & 0.25/0.42 \\
         seed(43) & 0.70/1.31 & 0.28/0.35 & 0.34/0.58 & 0.30/0.48 & 0.26/0.43\\
         seed(123) & 0.70/1.34	& 0.28/0.38	& 0.33/0.57	& 0.30/0.47 & 0.26/0.44 \\ 
         seed(222) & 0.67/1.29 & 0.32/0.42 & 0.35/0.53 & 0.29/0.46 & 0.26/0.42 \\
         \specialrule{0em}{1pt}{1pt}
         \hline
         \specialrule{0em}{1pt}{1pt}
         Avg. $\pm$ Std. &  \makecell{0.69$\pm$0.01/ \\ 1.27$\pm$0.07} & \makecell{0.30$\pm$0.02/\\0.43$\pm$0.06} & \makecell{0.33$\pm$0.02/\\0.56$\pm$0.03} & \makecell{0.30$\pm$0.01/\\0.49$\pm$0.03} & \makecell{0.26$\pm$0.00/\\0.43$\pm$0.01}\\
         \bottomrule
    \end{tabular}
    }
    \label{tab:random}
\end{table}

\subsection{Task Loss Function in GAN}
We evaluate the impact of different task loss functions in adversarial training on the predictive power of DTGAN and report the results as shown in Table \ref{tab:loss functions}. 
In two distributions, the Gaussian distribution is better than the uniform distribution, we conjecture the possible reasons: 1) Gaussian distribution is a common distribution in nature and has excellent mathematical properties. 2) Pedestrian trajectories are more likely to follow a Gaussian distribution rather than a uniform distribution.
It should be noted that DTGAN-M introducing MSE loss in adversarial training hardly improves compared to DTGAN, where we can see the loss function is a vital influence in the neural network optimization process.
Therefore, to improve the performance of the GAN-based framework in specific tasks, it is meaningful to explore reasonable task loss functions in adversarial training.

\subsection{Ablation Study and Component Analysis}
We conduct extensive ablation experiments in the Generator and Discriminator and validate them on all datasets to grasp the impact of the various components of the DTGAN model. \par

\textbf{Ablation Study}: In Table \ref{tab:ablation}, adding an SPE layer can improve the model performance (Variant 1 vs 3, 10 vs 11).
This suggests that embedding nodes into higher dimensions may benefit the model to get better trajectory representation.

\textbf{Component Analysis}: (\textit{Generator and Discriminator}) In Table \ref{tab:ablation}, GAT outperforms GCN (Variant 1 vs 5, 6 vs 10, 7 vs 9), GAT+CNN structure outperforms GCN+CNN (such as Social-STGCNN) structure, including other structures, this demonstrates the way in which pre-defined interactions limit the ability of neural networks, which is mentioned in the introduction.
Furthermore, CNN outperforms LSTM in the Generator (Variant 2 vs. 9, and 3 vs. 11), while LSTM outperforms CNN in the Discriminator (Variant 1 vs. 2, 6 vs. 7, and 9 vs. 10), which shows CNN can also handle time-series data very well in terms of forecasting, LSTM is slightly better than CNN in temporal data classification.

\begin{table}[t]
\renewcommand\arraystretch{1.5}
\centering
\caption{{\small Ablation study and Component analysis of DTGAN on the average ADE/FDE metrics.}}
\small
\resizebox{1.0\linewidth}{!}{
\begin{tabular}{c|ccccc|ccc|c}

\toprule[1pt]
{\large Variant}  & \multicolumn{5}{c|}{\large Generator} & \multicolumn{3}{c|}{\large Discriminator} & \multicolumn{1}{c}{\large Performance} \\
\Xcline{2-10}{0.5pt}
\large ID & GAT & GCN & LSTM & CNN & SPE & SPE & CNN & LSTM &
Avg. ADE/FDE \\

\midrule[1pt]
\midrule[1pt]
\large 1 & \checkmark & & & \checkmark & \checkmark & \checkmark & & \checkmark & 
{\Large 0.42/0.85} \\

\large 2 & \checkmark & & & \checkmark & \checkmark & \checkmark & \checkmark & &
{\Large 0.43/0.89}  \\

\large 3 & \checkmark & & & \checkmark & & \checkmark & & \checkmark & 
{\Large 0.44/0.94} \\

\large 4 & \checkmark & & & \checkmark & \checkmark & \checkmark & & & 
{\Large 0.45/0.90} \\

\large 5 & & \checkmark & & \checkmark & \checkmark & \checkmark & & \checkmark & 
 {\Large 0.46/0.94} \\
 
\large 6& & \checkmark & \checkmark & & \checkmark & \checkmark & & \checkmark &
{\Large 0.82/1.53} \\

\large 7& & \checkmark & \checkmark & & \checkmark & \checkmark & \checkmark & &
{\Large 0.85/1.65} \\

\large 8& & \checkmark & \checkmark & & & \checkmark & & \checkmark & 
{\Large 0.87/1.62} \\

\large 9& \checkmark & & \checkmark & & \checkmark & \checkmark & \checkmark & &
{\Large 0.87/1.59} \\

\large 10&\checkmark & & \checkmark & & \checkmark & \checkmark & & \checkmark & 
{\Large 0.90/1.63} \\

\large 11&\checkmark & & \checkmark & & & \checkmark & & \checkmark & 
{\Large 0.94/1.62} \\

\hline
\end{tabular}
}
\label{tab:ablation}
\end{table}

\subsection{Limitations}
Although DTGAN-G incorporates a log-likelihood function in adversarial training and achieves competative performance on ETH-UCY. Our work does not extensively explore its theoretical underpinning. We highly expect future work on that to delve into the study theoretically. In addition, the scope of our experimentation is limited to pedestrian trajectory datasets, diverse dataset types should be accounted for to extend the evaluation of DTGAN and DTGAN-G.

\section{Conclusion}
\label{section:conclusion}
In this work, we propose a novel framework, termed DTGAN, using random weights to free from pre-defined rules on the importance of graphs, which aims to draw implicit information from social interactions among all pedestrians. Besides, we explore various task loss functions in the adversarial training process of DTGAN to strike a balance between accuracy and realism among predictions. To the best of our knowledge, this is the first work that combines graph sequence data with random weights and adversarial training to extend the application of GAN.
The experimental study on the ETH and UCY datasets shows the effectiveness and precision of our framework.
The results demonstrate that our framework predicts pedestrian trajectories accurately, with superior achievements in direction prediction, stopping point prediction, and collision avoidance.  
In particular, the log-likelihood function introduced in adversarial training improves the performance of DTGAN. We expect to provide new directions of thinking and potential applications for GANs.
Furthermore, we assess the robustness of our framework to the influence of random weights input, validating its stability through statistical analyses. 
Moreover, through the comprehensive ablation study and component analyses, we gain valuable insights into the impact of each component of DTGAN, providing light on module design within the framework.
In the future, we will investigate the performance of DTGAN on diverse datasets like attribute-missing graphs and analyze the relevance between various components of the Generator and Discriminator.

\appendix
\begin{figure*}[htbp]
    \centering
    \begin{tabular}{cccc}
        \includegraphics[scale=0.17]{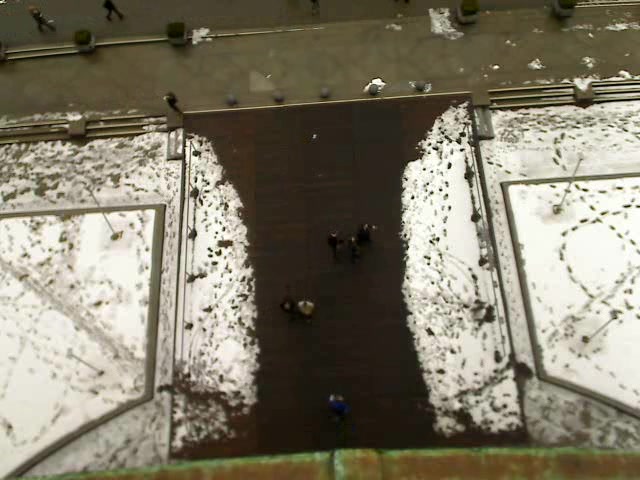} & \includegraphics[scale=0.14]{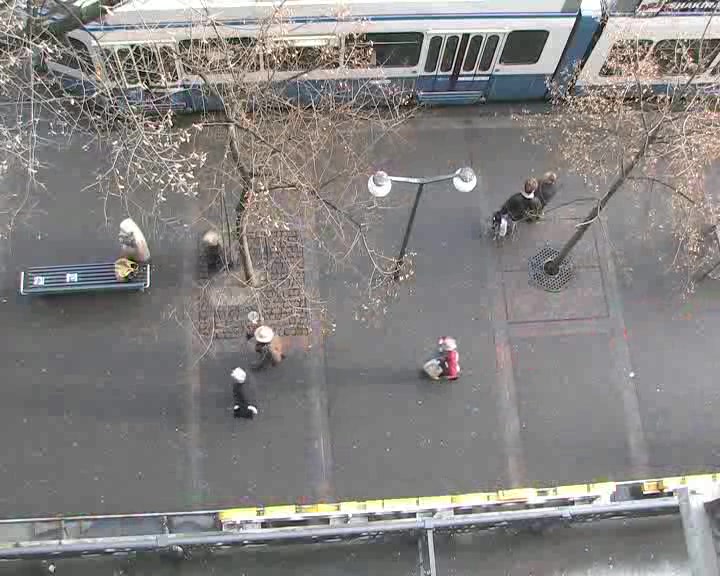} & \includegraphics[scale=0.14]{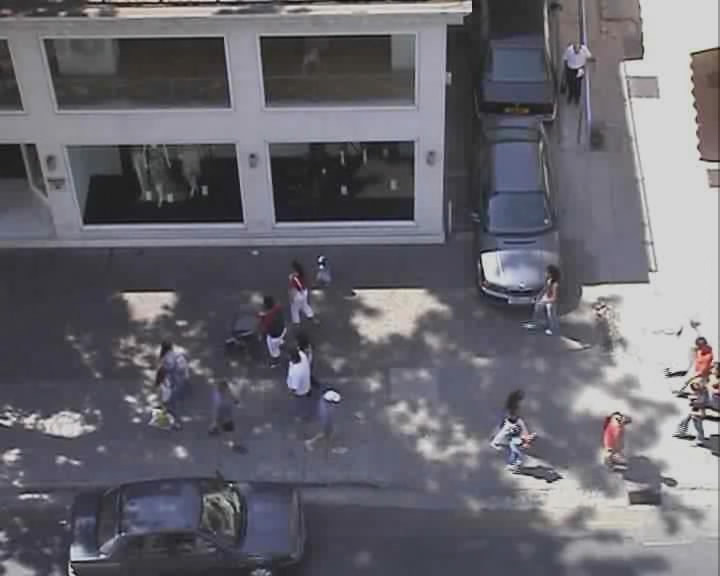} & \includegraphics[scale=0.14]{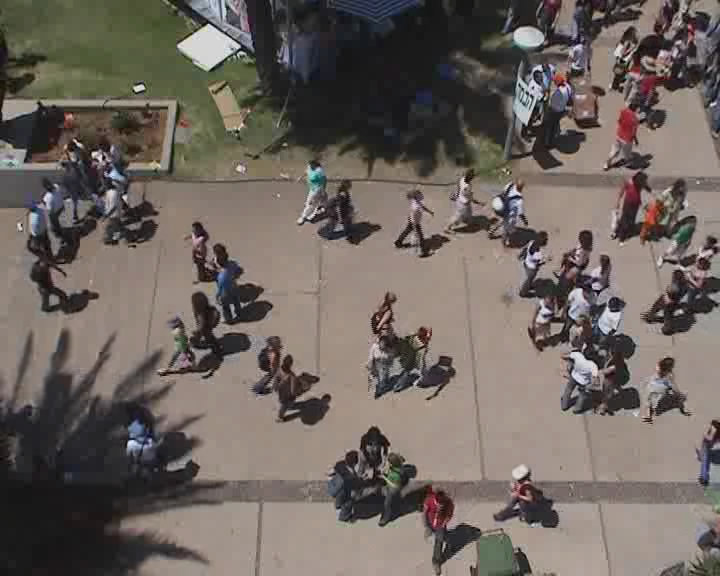} \\
        (a) ETH & (b) HOTEL & (c) ZARA & (d) UNIV \\
    \end{tabular}
    \captionsetup{justification=centering}
    \caption{Diverse scenarios from the ETH and UCY datasets illustrating different environments and crowd dynamics.}
    \label{fig:eth_ucy}
\end{figure*}
\section{appendix}
In this work, we use two public datasets ETH and UCY from real street surveillance videos. These data record the movement of pedestrians on the road from a bird's-eye view in different scenes in the form of videos. They are also widely used in the field of computer vision, such as target tracking, behavior analysis, and anomaly detection.

\textbf{ETH dataset }\footnote{https://icu.ee.ethz.ch/research/datsets.html}: there are two scenes, ETH and HOTEL, recording pedestrian movement patterns in front of the hotel and the light rail station. Figures \ref{fig:eth_ucy} (a) and (b) show two scenarios of the ETH dataset, respectively.
Among them, the ETH video records the scene overlooking the sidewalk from the top floor of the main building of the ``ETH Center'' on Sternworth Street. The video lasts about 8 minutes and marks about 750 pedestrians; the Hotel video records the scene overlooking the sidewalk from the fourth floor of a hotel on Bachhofst Street.  The video lasted about 13 minutes and marked about 750 different pedestrians.

\textbf{UCY dataset} \footnote{https://graphics.cs.ucy.ac.cy/research/downloads/crowd-data}: there are two scenes, ZARA and Univ, recording the movement patterns of pedestrians outside shopping malls and on university campuses and has thousands of nonlinear trajectories. Among them, the ZARA dataset records the scene of pedestrians passing through the door of the ``ZARA'' clothing store. The video lasts for 13 minutes, marking about 350 different pedestrians; Univ records the scene on a school road in the University of Cyprus. With a total video length of 7 minutes, approximately 620 different students were tagged.

\begin{algorithm}[htbp]
\SetKwInput{KwInput}{Input}
\SetKwInput{KwOutput}{Output}
\DontPrintSemicolon
    \KwInput{minPed, slen}
    \KwOutput{seqList}
    \KwData{D}
    seqList $:=$ [ ]\;
    
    time slot IDs $:=$ \textbf{Unique}(D[:,0])\;
    
    data $:=$ \textbf{Group}(D, time slot IDs)\;
    
    length $:=$ \textbf{Len}(time slot IDs) - slen + 1\;
    
    \For{i=0 to length}
    {
        seqtime slot $:=$ data[i:i+slen]\;
        
        pids $:=$ \textbf{Unique}(seqtime slot[:,1])\;
        
        seqPeds $:=$ \textbf{Group}(seqtime slot, pids)\;
        
        numPed $:=$ \textbf{Len}(pids)\;
        
        curSeq $:=$ []\;
        
        \For{$j=0$ \KwTo numPed}
        {
            seqPed $:=$ seqPeds[j]\;
            
            front $:=$ \textbf{Index}(time slot IDs, seqPed[0,0])\;
            
            end $:=$ \textbf{Index}(time slot IDs, seqPed[-1,0])\;
            
            
            \If{end-front+1 $\neq$ slen}
            {
                continue\;
            }
            \textbf{Append}(curSeq, seqPed)\;
        }
        
        \If{\textbf{Len}(curSeq) > minPed}
        {
            \textbf{Append}(seqList, curSeq)\;
        }
    }
    \caption{Extract multiple pedestrian trajectory data of specified time slot length}
    \label{alg:peds_seq}
\end{algorithm}

We represent our pedestrian trajectory dataset as $D=[d_1,d_2,\cdots,d_N]$, where each $d_i$ denotes the $i_{th}$ piece of data, comprising the time slot number, the pedestrian number, and pedestrian coordinates. The notation $D[:,0]$ extracts time slot IDs from all trajectories in $D$. We use the following method to process these raw datasets for training the generator. The detail is shown in the algorithm \ref{alg:peds_seq}, where \textit{Unique}, \textit{Group}, \textit{Len}, \textit{Index} and \textit{Append} are specific function functions. Their functions are to remove duplicate values, group by specified key values, obtain the sequence length, find the index of the specified value in the sequence, and add the value to the sequence. The output of this algorithm is a sequence of pedestrian trajectories with a specified time slot length sorted by time slot, which is defined as follows:
\begin{equation}
seqList = [seq^{(1)}_1, \cdots, seq^{(1)}_{m_1}, \cdots, seq^{(k)}_1, \cdots, seq^{(k)}_{m_k}],
\label{eq:seq_list}
\end{equation}

where $k$ represents the time slot number, $m_1,\cdots,m_k$ represents the number of pedestrians in each time slot. The number of pedestrians in different time slots is not necessarily the same. In this work, $\textit{minPed}$ is set to 3, that is, at least 3 pedestrians are needed to form a reasonable graph structure. According to the result of the formula \ref{eq:seq_list}, pedestrians with the same time slot number are classified into the same graph as nodes of the graph, thus forming graph sequence data:
\begin{equation}
seqGraph = [G^{(1)},\cdots,G^{(k)},\cdots],
\end{equation}
where $G^{(i)}=(V^{(i)}, E^{(i)})$. $V^{(i)}$ represents the nodes of the graph in time slot $i$, and each node attribute is the coordinate of the pedestrian. $E^{(i)}$ represents the edges of the graph.

We first extract multiple pedestrian trajectory data of a specified time slot length from the input data, store the data in a list, and return it. It mainly includes grouping the input data by the number of time slots and extracting a sequence of pedestrian trajectories with a number of consecutive time slots $slen$ from each group of time slot data. If the number of pedestrians contained in the sequence is greater than $\textit{minPed}$, it is stored in $\textit{seqList}$.

During the entire data processing process, the input original data needs to be traversed and calculated multiple times, and the original data is organized into a data set containing multiple sequences, where each sequence contains the position information and displacement information of multiple pedestrians in each time slot, process and normalize the data of each sequence, and finally divide all sequence data sets into training sets, verification sets, and test sets.
Therefore, we have provided a summary table \ref{tab1} of the number of samples used in training, validation, and testing for each dataset. To further elaborate:
Each dataset presents unique characteristics in terms of pedestrian behavior and environmental settings.
\begin{table}[h]
    \centering
    \caption{number of samples used of train, validation, and test in five sub-datasets}
    \begin{tabular}{c|c|c|c|c|c}
        \toprule
          & ETH & HOTEL & UNIV & ZARA1 & ZARA2 \\
          \hline
            train & 2785 & 2594& 2076& 2322 & 2112 \\
            validation &660 & 621 & 530 & 605 &501 \\
            test  & 70 & 301 & 947 & 602 & 921\\
         \hline

    \end{tabular}
        \label{tab1}
\end{table}

\bibliographystyle{IEEEtran}
\bibliography{ref}
\begin{IEEEbiography}[{\includegraphics[width=1in,height=1.25in,clip,keepaspectratio]{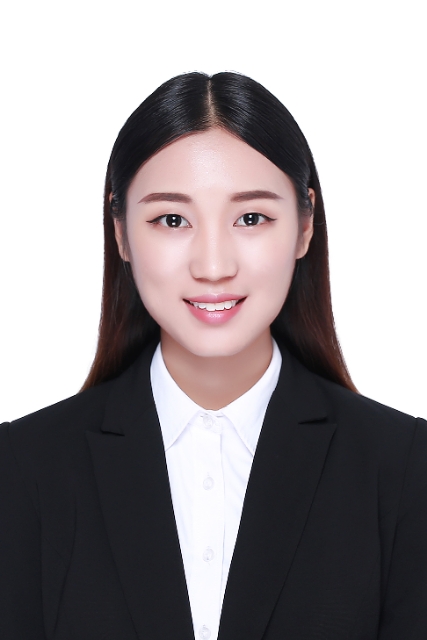}}]{Jiajia Xie} received the M.S. degree from the College of Computer Science and Electronic Engineering, Hunan University, Changsha, China. She was a Research Assistant/Algorithm Engineer at Zhejiang Lab, involved in the Traffic Big Data Mining and Analysis Project and Large Scale Efficient Graph Learning Platform Project. Her research interests include  Spatial-Temporal Data Mining, Deep Learning, and Optimization.
\end{IEEEbiography}
\begin{IEEEbiography}[{\includegraphics[width=1in,height=1.25in,clip,keepaspectratio]{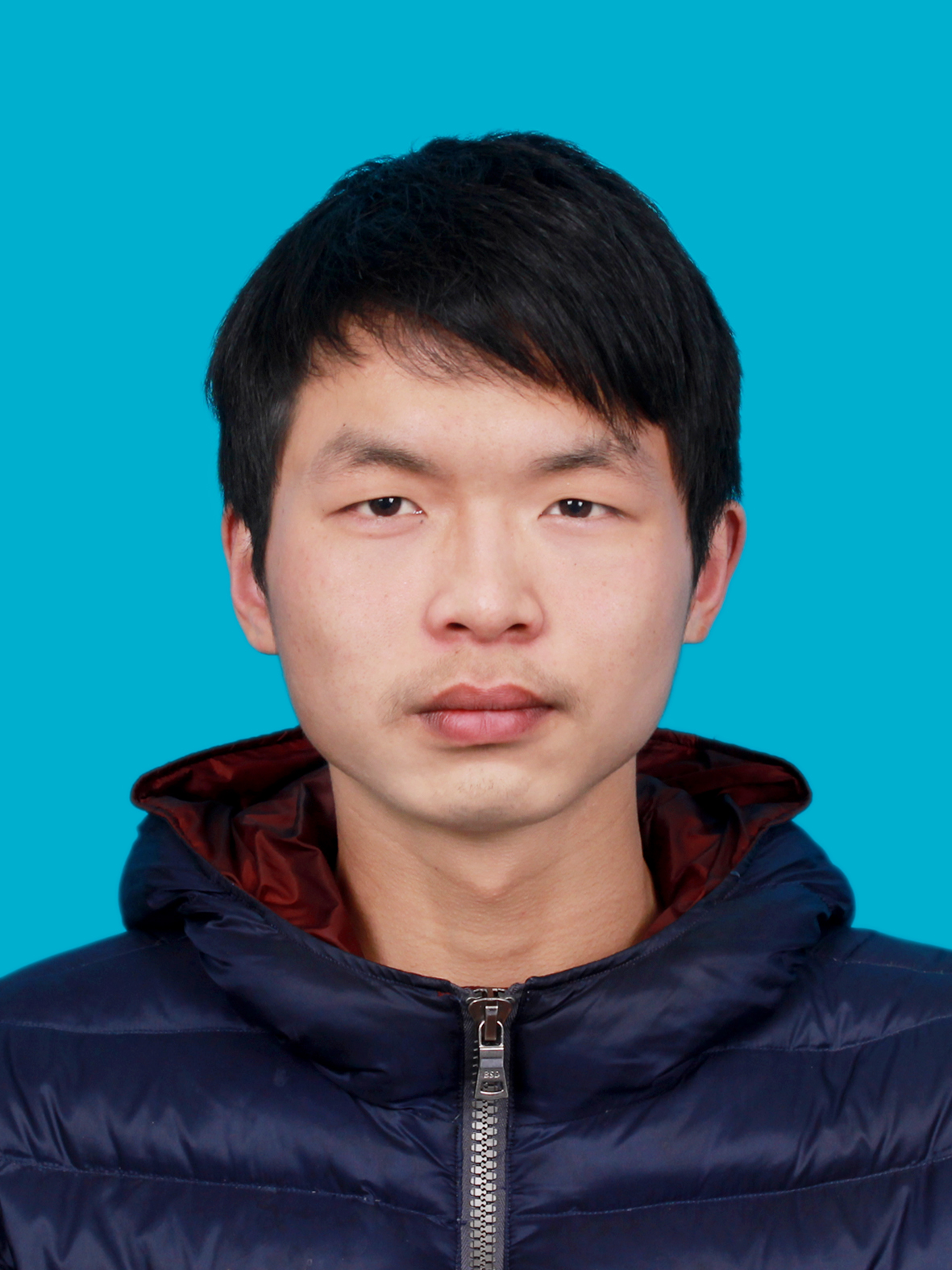}}]{Sheng Zhang} received the M.S. degree from the College of Computer Science and Technology in Zhejiang University. He is currently working in Zhejiang Lab, Hangzhou as an Algorithm Engineer. His research interests include graph neural networks, machine learning, and large language model training and acceleration.
\end{IEEEbiography}
\begin{IEEEbiography}[{\includegraphics[width=1in,height=1.25in,clip,keepaspectratio]{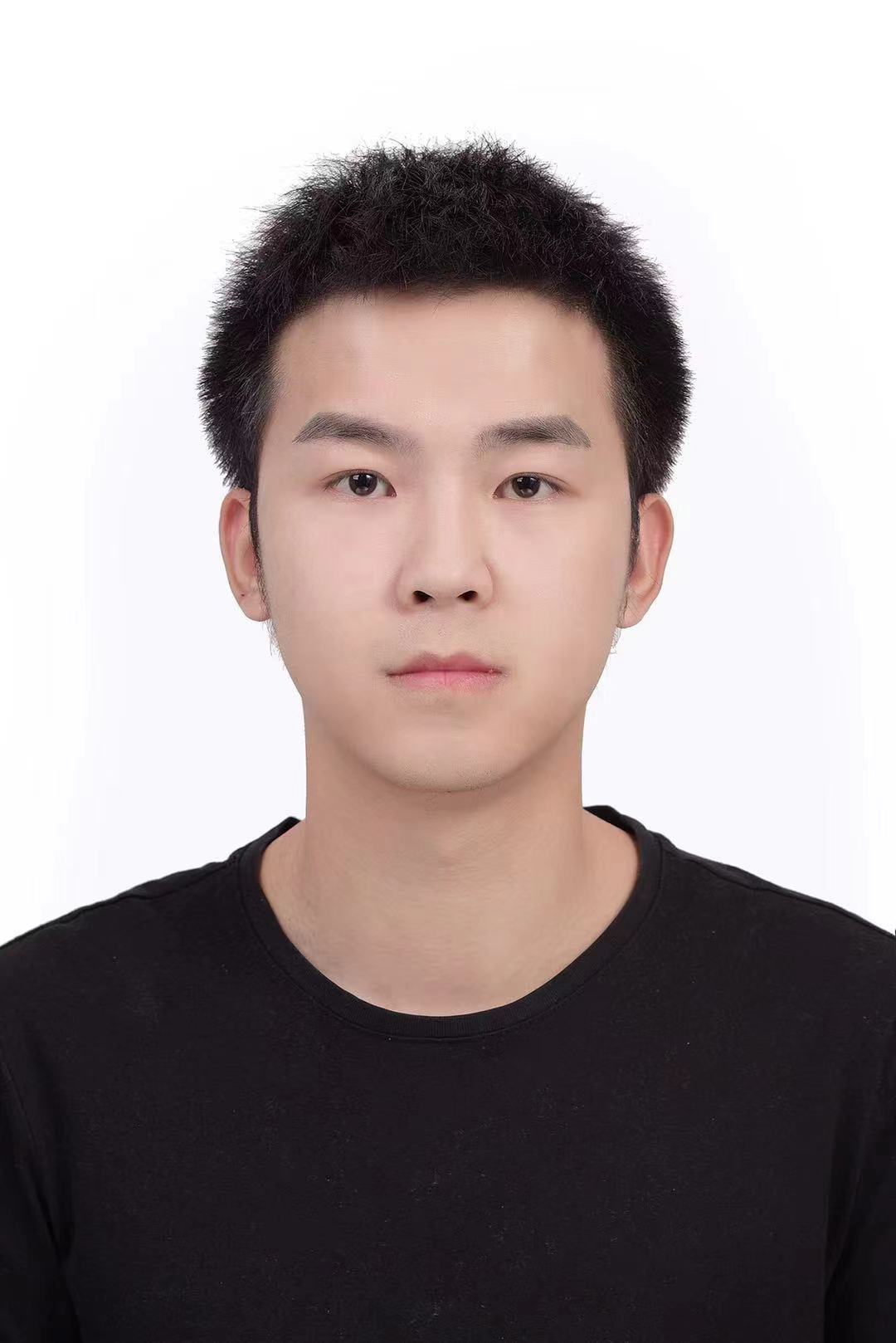}}]{Beihao Xia} received the B.Eng degree from the College of Computer Science and Electronic Engineering, Hunan University, Changsha, China, in 2015, the M.S. and Ph.D. degrees from the School of Electronic Information and Communications, Huazhong University of Science and Technology, Wuhan, China, in 2018 and 2023, respectively. His current research interests include trajectory prediction, behavior analysis and understanding.
\end{IEEEbiography}

\begin{IEEEbiography}[{\includegraphics[width=1in,height=1.25in,clip,keepaspectratio]{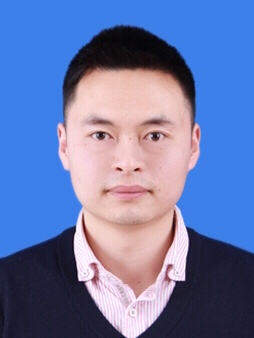}}]{Zhu Xiao} (M’15-SM’19) received the M.S. and Ph.D. degrees in communication and information
 systems from Xidian University, China, in 2007 and
 2009, respectively. From 2010 to 2012, he was a
 Research Fellow with the Department of Computer
 Science and Technology, University of Bedfordshire,
 U.K. He is currently a Full Professor with the College
 of Computer Science and Electronic Engineering,
 Hunan University, China. His research interests include wireless localization, Internet of Vehicles and intelligent transportation systems. He is a senior member of the IEEE. He
 is currently an Associate Editor for the IEEE Transactions on Intelligent Transportation Systems.
\end{IEEEbiography}

\begin{IEEEbiography}[{\includegraphics[width=1in,height=1.25in,clip,keepaspectratio]{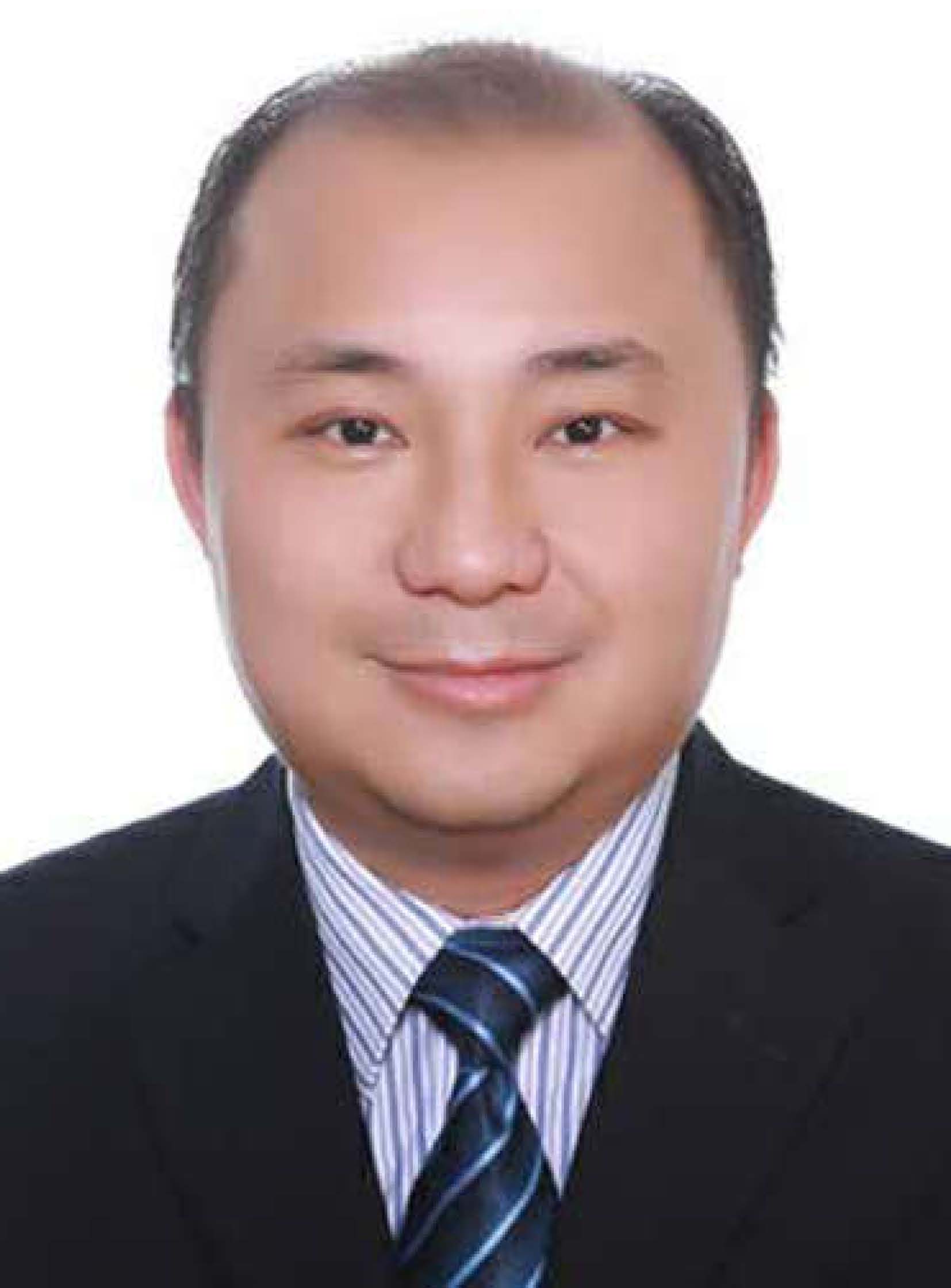}}]{Hongbo Jiang}
(Senior Member, IEEE) received the
 Ph.D. degree from Case Western Reserve University
 in 2008. He was a Professor with the Huazhong
 University of Science and Technology. He is currently a Full Professor with the College of Computer Science and Electronic Engineering, Hunan University. His research interests include computer networking, especially algorithms, and protocols for
 wireless and mobile networks. He is an Elected
 Member of Academia Europaea, a fellow of IET,
 a fellow of BCS, and a fellow of AAIA. He was the
 Editor of IEEE/ ACM Transactions on Networking, the Associate Editor of IEEE Transactions on Mobile Computing, and the Associate Technical Editor of IEEE Communications Magazine.

\end{IEEEbiography}

\begin{IEEEbiography}[{\includegraphics[width=1in,height=1.25in,clip,keepaspectratio]{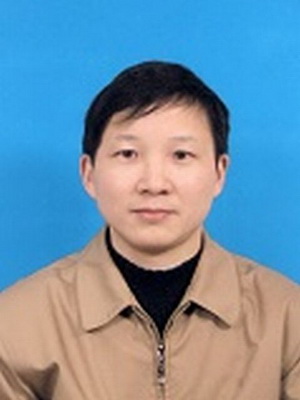}}]{Siwang Zhou}
received the B.S. degree from Fudan University, Shanghai, China, the M.S. degree from Xiangtan University, Xiangtan,
China, and the Ph.D. degree from Hunan University,
Changsha, China. He is currently a Professor with
the College of Computer Science and Electronic Engineering,
Hunan University. His research interests
include image compressive sensing, deep learning,
and Internet of Things.
\end{IEEEbiography}

\begin{IEEEbiography}[{\includegraphics[width=1in,height=1.25in,clip,keepaspectratio]{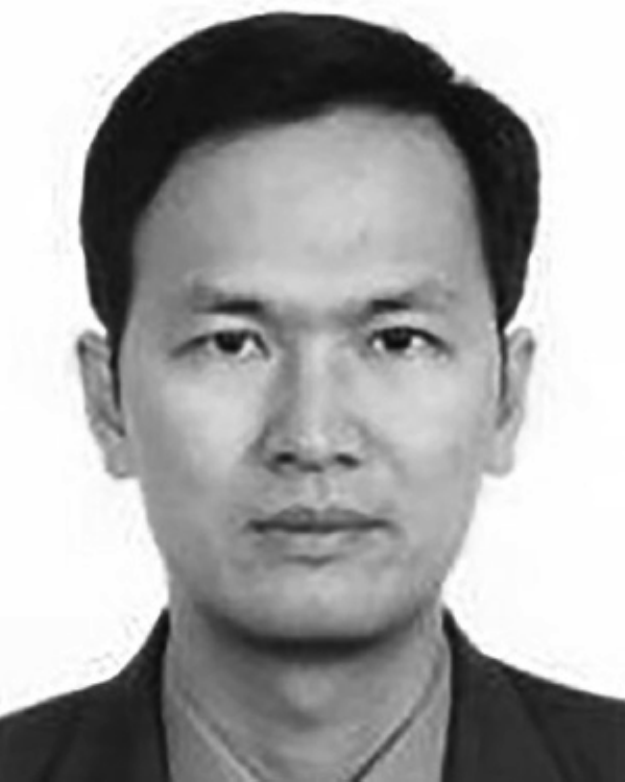}}]{ Zheng Qin}
 received the Ph.D. degree in computing
 science from Chongqing University in 2001. He is
 a Professor and the Vice President of the College
 of Computer Science and Electronic Engineering,
 Hunan University. He is the Deputy Director of the
 Engineering Laboratory of Cyberspace Identity and
 Data Security in Hunan Province, a special expert
 of the Ministry of National Security, and a member
 of the CCF Big Data Special Committee. His main
 research interests are big data/cloud computing, net
work information security and privacy protection,
 smart city, and large complex software engineering.
\end{IEEEbiography}

\begin{IEEEbiography}[{\includegraphics[width=1in,height=1.25in,clip,keepaspectratio]{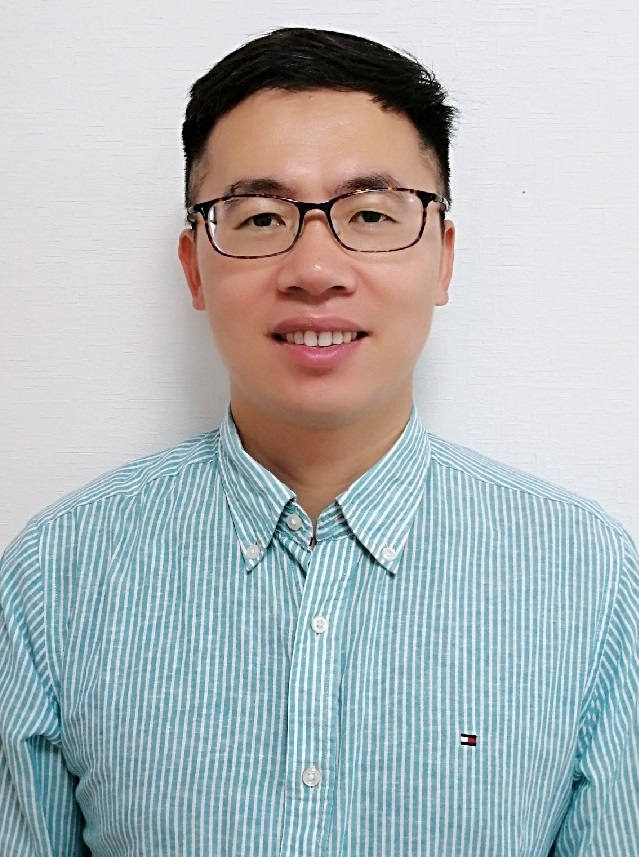}}]{Hongyang Chen}
received the B.S. degree from Southwest Jiaotong University, Chengdu, China, in 2003, the M.S. degree from
the Institute of Mobile Communications, Southwest
Jiaotong University, in 2006, and the Ph.D. degree
from The University of Tokyo in 2011. From 2004 to 2006, he was a Research Assistant with the Institute of Computing Technology, Chinese Academy of Sciences. In 2009, he was a Visiting Researcher with the UCLA Adaptive Systems Laboratory, University of California Los Angeles.
From April 2011 to June 2020, he worked as a Researcher with Fujitsu Ltd., Japan. He is currently a Principal Investigator/Senior Research Scientist with the Zhejiang Laboratory, China. He has authored or coauthored more
than 80 refereed journals and conference papers. He has granted/filed more than 50 PCT patents. His research interests include data-driven intelligent networking and systems, machine learning, localization and location-based
big data, B5G, and statistical signal processing. He was the Editor of IEEE TRANSACTIONS ON WIRELESS COMMUNICATIONS and the Associate Editor of IEEE COMMUNICATIONS LETTERS.
\end{IEEEbiography}

\end{document}